\documentclass[10pt,twocolumn,letterpaper]{article}

\usepackage{cvpr}              %

\usepackage{multirow}
\usepackage{booktabs}  %
\usepackage{multirow}  %
\usepackage{amsmath}   %
\usepackage[normalem]{ulem}
\usepackage{float}
\usepackage{siunitx}
\usepackage{adjustbox}
\usepackage{caption}

\definecolor{cvprblue}{rgb}{0.21,0.49,0.74}
\usepackage[pagebackref,breaklinks,colorlinks,allcolors=cvprblue]{hyperref}

\def\paperID{12224} %
\def\confName{CVPR}
\def\confYear{2025}

\title{Uni4D: Unifying Visual Foundation Models for %
4D Modeling from a Single Video}

\author{
David Yifan Yao
~~~~~~~~
Albert J. Zhai
~~~~~~~~
Shenlong Wang\\
University of Illinois at Urbana-Champaign
}

\begin{document}

\twocolumn[{
\renewcommand\twocolumn[1][]{#1}
\maketitle
\begin{center}
    \vspace{-10mm}
    \includegraphics[width=\textwidth]{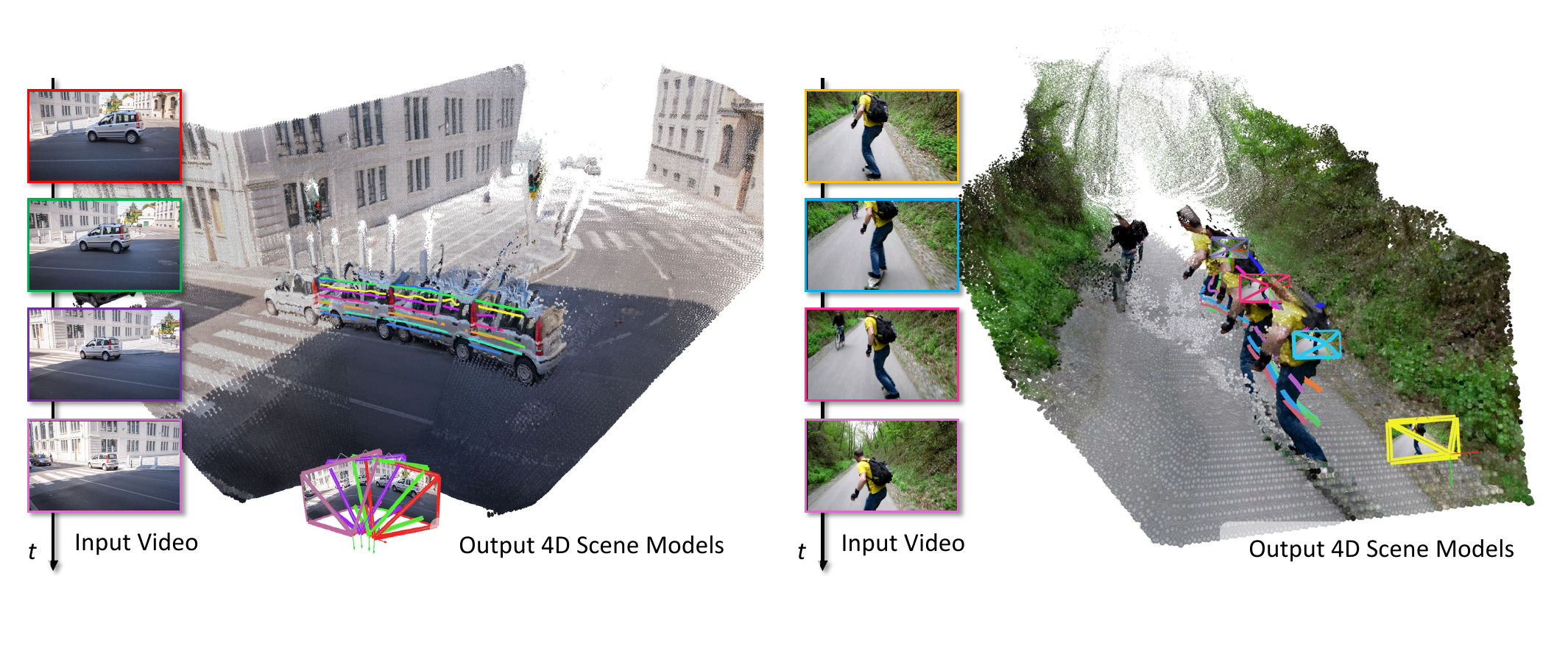}
    \vspace{-15mm}
    \captionof{figure}{Given a casually captured video, {\bf Uni4D} harnesses pretrained visual foundation models and multi-stage optimization to jointly estimate camera poses, dynamic geometry, and dense 3D motion. The resulting camera poses and geometry are accurate, consistent, and coherent both temporally and spatially. This is all done without any additional training or fine-tuning.}
    \label{fig:teaser}
\end{center}

}]

\begin{abstract}
This paper presents a unified approach to understanding dynamic scenes from casual videos. Large pretrained vision foundation models, such as vision-language, video depth prediction, motion tracking, and segmentation models, offer promising capabilities. However, training a single model for comprehensive 4D understanding remains challenging. We introduce Uni4D, a multi-stage optimization framework that harnesses multiple pretrained models to advance dynamic 3D modeling, including static/dynamic reconstruction, camera pose estimation, and dense 3D motion tracking. Our results show state-of-the-art performance in dynamic 4D modeling with superior visual quality. Notably, Uni4D requires no retraining or fine-
tuning, highlighting the effectiveness of repurposing visual foundation models for 4D understanding. Code and more results are available at: \href{https://davidyao99.github.io/uni4d/}{https://davidyao99.github.io/uni4d}.

\end{abstract}
    
\section{Introduction}
\label{sec:intro}

Over the past two years, many visual foundation models have emerged~\cite{piccinelli2024unidepth, yang2024depth, ke2024repurposing, kirillov2023segment, khirodkar2025sapiens, bae2024rethinking, wang2024dust3r, karaev2023cotracker, le2024dense}, achieving high accuracy on tasks like depth prediction, segmentation, human parsing, normal estimation, few-view reconstruction, and motion tracking. These models leverage supervised learning on large, diverse datasets, achieving impressive accuracy and remarkable generalization. However, these advances have not translated to 4D (time + geometry) modeling, a longstanding challenge in computer vision. We see two main reasons: first, collecting high-quality, ground-truth 4D data from real-world environments remains complex and resource-intensive. Second, 4D understanding is a holistic problem involving interconnected tasks like camera pose estimation, 3D reconstruction, and dynamics tracking. Although each subtask shows progress, data-driven cues remain noisy, and more importantly, unifying them synergistically for holistic modeling remains challenging. This paper seeks to answer: {\it Can we harness the success of visual foundation models for dynamic 4D modeling?}

In this paper, we propose Uni4D, a novel framework that reconstructs 4D scenes from a single video captured in the wild. Our method integrates data-driven foundation models and conventional model-driven dynamic structure-from-motion, combining data-driven cues and model-based knowledge synergistically. Our intuition is simple: each data-driven visual cue, such as video segmentation, pixel-level motion tracking, and video depth, is a partial projection from the 4D world to the 2D video. The key is to create a 4D scene representation that coherently aligns with each cue while incorporating strong prior knowledge of real-world motion and shape to resolve temporary inconsistencies. To achieve this, we take an energy-minimization perspective, framing the problem as an optimization task that jointly infers camera poses, static and dynamic geometry, and motion.
The dynamic 4D modeling framework is complex, as it involves various optimization variables, visual cues, and constraints. To overcome the challenge of joint reasoning, we carefully design a novel three-stage, divide-and-conquer pipeline that progressively incorporates camera poses, static geometry, and dynamic geometry and motion into the optimization framework.
Uni4D leverages pretrained foundation models across different tasks, requiring no task-specific retraining or fine-tuning. This design sidesteps the need for 4D ground-truth data, a major challenge in the field. Through incorporating strong priors on geometry and dynamic motion, our method produces realistic 4D scenes that are coherent across space and time while maintaining high accuracy.

\begin{figure}[ht] %
    \centering
    \vspace{-5mm}
\includegraphics[width=0.7\linewidth]{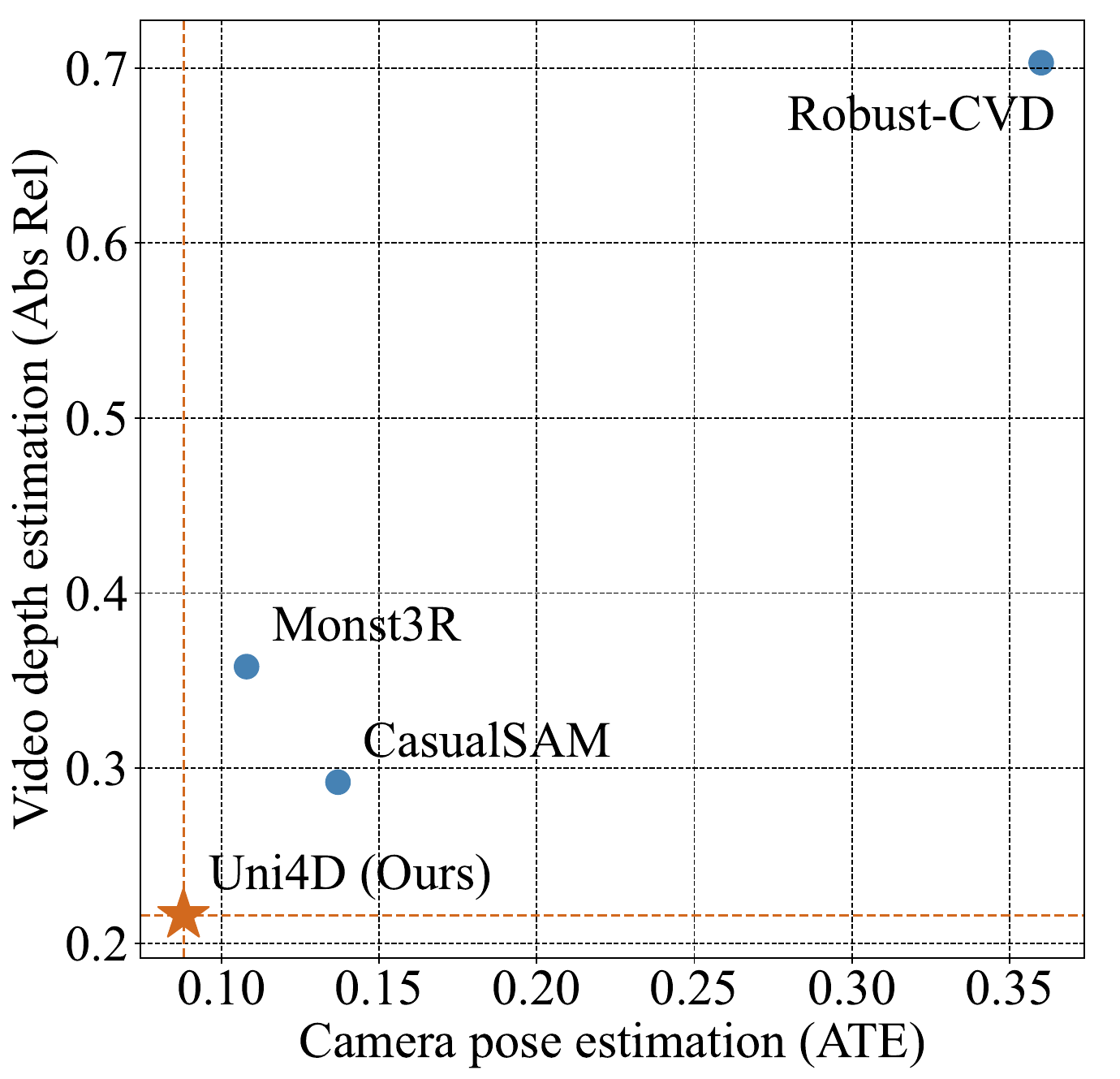}
    \vspace{-3mm}
    \caption{{\bf Uni4D} outperforms other recent 4D modeling methods in both camera pose and geometry accuracy on the Sintel dataset.}
    \label{fig:keyresult}
    \vspace{-5mm}
\end{figure}

We demonstrate the effectiveness of Uni4D on various datasets. As shown in Fig.~\ref{fig:teaser}, our method effectively recovers the clean geometry and motion of the scene, as well as camera trajectories, from a single video. As shown in Fig.~\ref{fig:keyresult}, our model outperforms all dynamic 4D modeling baselines in both camera pose and geometry.

\section{Related Works}
\label{sec:related}

\paragraph{Structure from Motion and SLAM.} Structure from Motion (SfM) is a classical problem in computer vision that aims to jointly recover camera parameters and 3D structure from images~\cite{snavely2008modeling, agarwal2011building, schonberger2016structure, cui2017hsfm, pan2024global}. Simultaneous localization and mapping (SLAM) is a similar problem that focuses on real-time efficiency in an online setting~\cite{davison2007monoslam, engel2014lsd, mur2015orb, teed2021droid}. The core idea for many approaches to SfM and SLAM is to optimize a form of reprojection error with respect to both camera parameters and 3D points, also known as bundle adjustment~\cite{triggs2000bundle}. This relies on the key assumption that the scene being viewed is rigid (``static''), making these approaches not applicable to dynamic scenes. 

\paragraph{4D Understanding.} 
The problem of non-rigid structure from motion is highly ill-posed. Previous works on non-rigid SfM leveraged various forms of priors or weaker assumptions on shape and or motion to make the problem solvable~\cite{bregler2000recovering, zhu2014complex, kong2016prior, kong2016structure, agudo2018image}. With the rise of modern segmentation models, many works have studied category-specific articulation and rigidity priors, focusing on object categories such as humans~\cite{sun2023trace, ye2023slahmr, goel2023humans, song2023total}, animals~\cite{yang2022banmo, wu2022casa, zhang2023slomo, yang2023ppr}, and vehicles~\cite{ma2019deep, gojcic2021weakly}. The generality of such methods is limited by their category-specific nature.
Recently, several works have also begun to leverage trained neural networks as priors for 4D understanding in a more freeform manner~\cite{kopf2021robust, zhao2022particlesfm, zhang2022structure, zhang2024monst3r}. However, these methods involve either training a network~\cite{zhao2022particlesfm, zhang2024monst3r} or optimizing specific layers of an existing network~\cite{kopf2021robust, zhang2022structure}, making it difficult to integrate newer models into the pipeline. Our method takes the strategy further and integrates pretrained models in a completely modular manner, fully unleashing their generalization capabilities and allowing for seamless integration of newer models. Note that various recent works have also studied 4D reconstruction in an optimization-based framework~\cite{gao2021dynamic, li2023dynibar, wu20244d, wang2024shape, lei2024mosca}. However, these methods have additional rendering capacities and focus on rendering metrics, while we focus on recovering high-quality geometry.

\paragraph{Visual Foundation} Models. In recent years, a number of visual foundation models have been developed, achieving remarkable performance on tasks like depth estimation~\cite{piccinelli2024unidepth, ke2024repurposing, yang2024depth, hu2024depthcrafter, bochkovskii2024depth}, detection and segmentation~\cite{liu2023grounding, kirillov2023segment, ravi2024sam, deitke2024molmo}, human parsing~\cite{khirodkar2025sapiens}, surface normal estimation~\cite{bae2024rethinking, fu2025geowizard}, few-view reconstruction~\cite{wang2024dust3r, leroy2024grounding}, and point tracking~\cite{karaev2023cotracker, karaev2024cotracker3, le2024dense}. 
Our insight is that nearly all of these models have the potential to contribute towards holistic 4D understanding, and harnessing them in a unifying framework can advance the state-of-the-art in tasks such as non-rigid structure from motion. To this end, we integrate the following pretrained models in Uni4D: UniDepthv2~\cite{piccinelli2025unidepthv2} for geometry initialization, CoTracker3~\cite{karaev2024cotracker3} for correspondence initialization, and a collection of Recognize Anything Model~\cite{zhang2024recognize}, ChatGPT~\cite{achiam2023gpt}, Grounding-SAM~\cite{liu2023grounding, kirillov2023segment}, and DEVA~\cite{cheng2023tracking} for dynamic object segmentation. Through multi-stage optimization with a few regularizing priors, we are able to use these models to extract accurate pose and 4D geometry from monocular video.

\section{Method}
\label{sec:method}

\begin{figure*}[htbp]
    \vspace{-5mm}
    \centering
    \includegraphics[width=.9\textwidth]{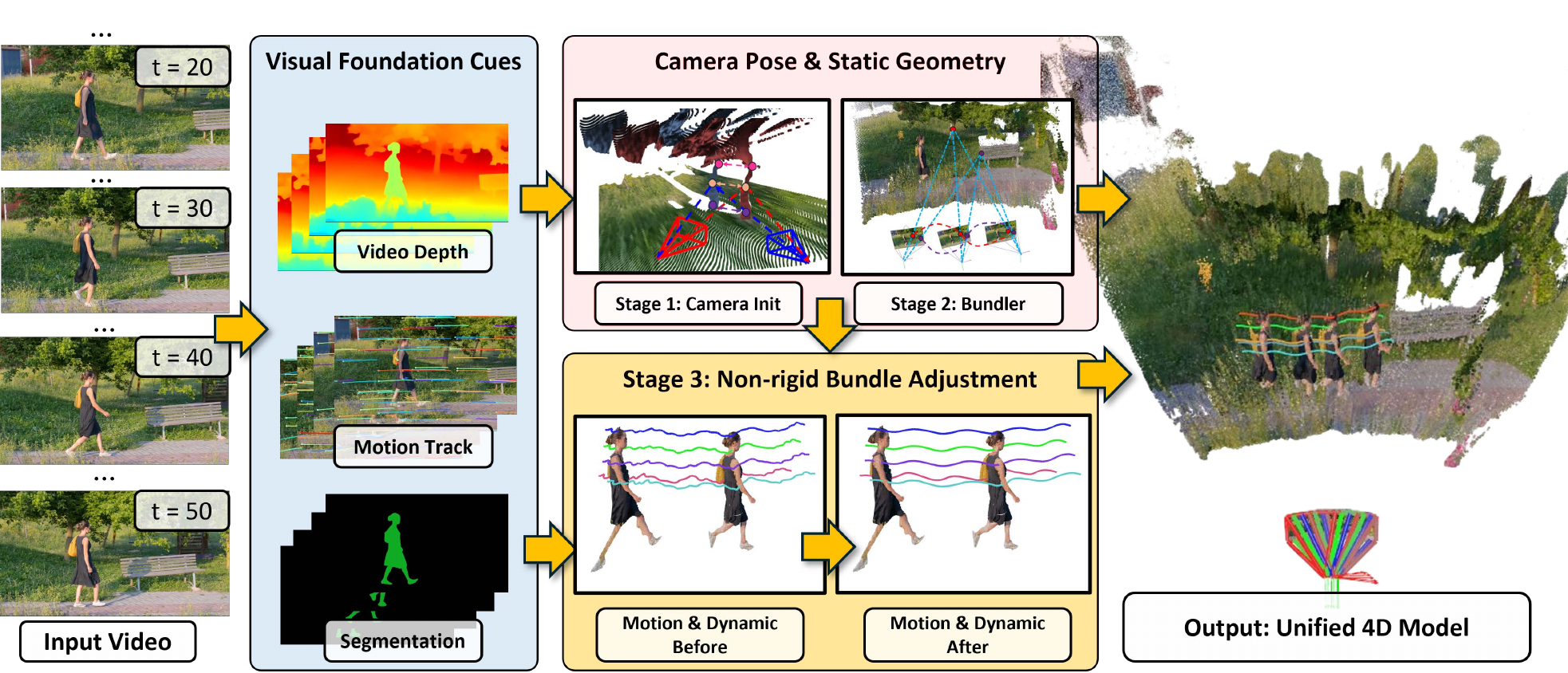}
    \vspace{-2mm}
    \captionof{figure}{Given a casually captured video, {\bf Uni4D} exploits visual foundation models to extract dynamic segmentation, video depth, and motion tracks. Static geometry and poses are obtained through tracklet-based structure-from-motion along with camera motion priors. Dynamic geometry is improved through nonrigid bundle adjustment and scene motion priors. A final fusion densifies geometry to obtain high quality 4D reconstruction.}
    \vspace{-5mm}
    \label{fig:method}
\end{figure*}

In this paper, we are interested in recovering 4D geometry and camera parameters from a monocular casual video.
Our model is built on the intuition that the 2D visual cues from the video can be seen as perspective projections of 4D geometry and motion, where the video depth represents a projection of 4D geometry, 2D dense tracking corresponds to a projection of 4D motion, and segmentation reflects a projection of the 4D dynamic object silhouette. 
We propose a novel, training-free, foundation-model-based energy minimization scheme that leverages visual cues to jointly infer camera poses, geometry, and motion, enabling generalization across diverse videos.

\subsection{Pretrained Visual Cues}
\label{subsec:cues}

We first exploit foundation models to extract visual cues, including dynamic segmentation, video depth, and motion tracking. All of the models are pre-trained on large and diverse datasets and exhibit strong generalization abilities.  

\paragraph{Video Segmentation.} 
Recognizing and segmenting dynamic objects is crucial for 4D scene understanding. We leverage the latest advancements in video semantic segmentation and tracking to estimate dynamic objects over time. First, we use RAM~\cite{zhang2024recognize} to identify semantic classes in the video. These classes are then filtered through GPT-4o~\cite{achiam2023gpt} to exclude static and background elements (e.g., buildings, poles), retaining only dynamic objects (e.g., humans, animals, vehicles). Next, Grounding-SAM~\cite{liu2023grounding, kirillov2023segment} performs segmentation at each keyframe, and DEVA~\cite{cheng2023tracking} tracks these segments over time, resulting in accurate dynamic video segmentation~$\{ \mathbf{M_t}\}_{t=0}^T$.

\begin{figure*}[htbp]    %
    \centering
    \vspace{-5mm}
    \includegraphics[width=.8\textwidth]{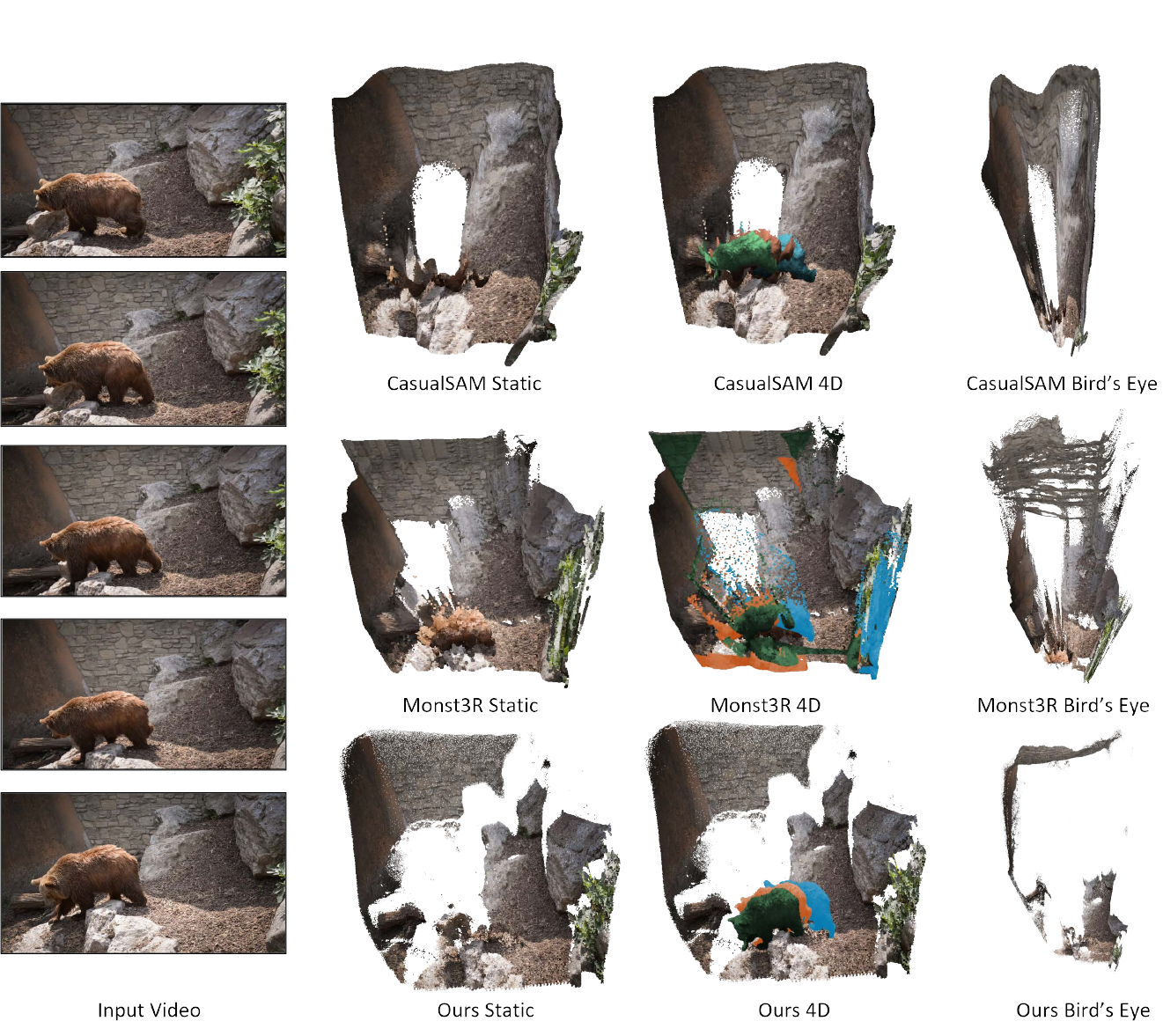}
    \vspace{-3mm}
    \caption{{\bf Qualitative results of 4D reconstruction on the DAVIS~\cite{perazzi2016benchmark} dataset.} CasualSAM~\cite{zhang2022structure} suffers from slanted geometry, and Monst3R~\cite{zhang2024monst3r} has unclear geometry and does not resolve conflicts from multiple views (note the wall in the bird's-eye view). Both CasualSAM and Monst3R lack clean dynamic reconstruction and segmentation. Uni4D achieves a realistic layout, thanks to joint optimization, and provides accurate dynamic segmentation and reconstruction by leveraging foundation visual models as cues. %
    }
    \vspace{-3mm}
    \label{fig:qual_davis_bear}
\end{figure*}
\paragraph{Dense Motion Tracking.} 
We use dense pixel tracking to establish correspondences over time, which serve as motion cues to assist inference of geometry reconstruction and dynamic object 3D motion. Unlike traditional 4D reconstruction methods relying on optical flow~\cite{zhang2024monst3r, zhang2022structure}, dense motion tracklets yield more correspondence pairs across large viewpoints and structural changes. 
Pixel tracking also outperforms sparse matching in density and surpasses flow propagation in robustness, making it ideal for dynamic 4D reconstruction.%
We utilize Co-TrackerV3~\cite{karaev2024cotracker3} for its robustness, as it employs a 4D cost volume with an attention mechanism to track 2D points through occlusions, recently proving to be effective for dynamic neural rendering. We apply Co-Tracker bi-directionally 
on a dense grid every 10 frames to ensure thorough coverage. %
We filter and classify tracklets using segmentation masks yielding a set of correspondent point trajectories $\{ \mathbf{Z}_{k} \in \mathbb{R}^{T\times 2} \}_{k=0}^K$ at visible time steps determined by Co-Tracker.

\paragraph{Video Depth Estimation.} Monocular depth reasoning, while insufficient as a standalone tool for complete 4D geometry recovery, provides strong geometry initialization cues. In our model, we use UniDepthV2~\cite{piccinelli2025unidepthv2}, a monocular depth estimation network, to estimate an initial depth map, $\{ \mathbf{D_t}\}_{t=0}^T$, and initial camera intrinsics, $\mathbf{K_{init}}$.

\subsection{Energy Formulation}
\label{subsec:formulation}

We now describe the energy formulation of our proposed dynamic 4D reconstruction framework. 
Let $\mathcal{M} = \{\mathbf{M}_t\}_{t=0}^T$, $\mathcal{D} = \{\mathbf{D}_t\}_{t=0}^T$, $\mathcal{Z} = \{\mathbf{Z}_k\}_{k=0}^K$ be the input dynamic segmentation, monocular depth, dense motion trajectory extracted from the input video $\mathcal{I} = 
\{\mathbf{I_t}\}_{t=0}^T$.
Formally, our goal is to obtain camera parameters $\mathcal{C}$, namely poses $\mathcal{T}$ and intrinsics $\mathbf{K}$, and a set of 4D point clouds $\mathcal{P}$ containing both dynamic and static parts separately as $\mathcal{P} = \{ \mathbf{P}_\mathrm{static}, \mathcal{P}_\mathrm{dyn}\}$.
Here, $\mathbf{P}_\mathrm{static}$ does not change over time and $\mathcal{P}_\mathrm{dyn} = \{\mathbf{p}_k \in \mathbb{R}^{T\times 3}\}_{k}$ is a temporally varying point cloud, where each $\mathbf{p}_k$ is a dynamic point trajectory.  
We represent the camera poses as rigid transforms $\mathcal{T} = \{ \boldsymbol{\xi}_t \in \mathbb{SE}(3)\}_{t=0}^T$, and parameterize the rotations as $\mathfrak{so}(3)$ rotation vectors as a minimal representation for easy optimization. We assume all frames share the same intrinsic matrix $\mathbf{K}$ where we optimize focal lengths $f_x$ and $f_y$.  We formulate the 4D joint reasoning problem as minimization of the following energy function:
\vspace{-0.3mm}
\begin{equation}
\label{eq:total}
E_\mathrm{BA}(\mathcal{C}, \mathbf{P}_\mathrm{static})+  
E_\mathrm{NR}(\mathcal{P}_\mathrm{dyn}) + 
E_\mathrm{motion}(\mathcal{P}_\mathrm{dyn}) + 
E_\mathrm{cam}(\mathcal{T}) 
\end{equation}
where $E_\mathrm{BA}(\mathcal{C}, \mathbf{P}_\mathrm{static})$ is a bundle adjustment term that measures the discrepancy between static-region correspondences and the static 3D structure $\mathbf{P}_\mathrm{static}$ through perspective reprojection. $E_\mathrm{NR}(\mathcal{P}_\mathrm{dyn})$ is a non-rigid structure-from-motion energy term that measures the disagreement between the dynamic point cloud and their tracklet correspondences, $E_\mathrm{cam}(\mathcal{T})$ is a regularization term on the camera motion smoothness, and $E_\mathrm{motion}(\mathcal{P}_\mathrm{dyn})$ is a regularization term on the dynamic structure and motion. Each energy term is involved in different stages of the optimization process, which will be described in Sec.~\ref{sec:inference}.

\paragraph{Static Bundle Adjustment Term.} The bundle adjustment energy $E_\mathrm{BA}(\mathcal{C}, \mathbf{P}_\mathrm{static})$ measures the consistency between the pixel-level correspondences and the 3D structure of \textbf{static} scene elements. Given the input pixel tracks $\mathcal{Z}=\{ \mathbf{Z}_k \}$ and video segmentation $\mathcal{M}$, we filter all tracks corresponding to static areas and 
minimize the distance between the projected pixel location and the observed pixel location: 

\begin{equation}
E_\mathrm{BA}(\mathcal{C}, \mathbf{P}_\mathrm{static};\mathcal{Z}, \mathcal{M})
=\sum_{\mathbf{z}_{k} \in \mathcal{M}} \sum_t \mathbf{w_{k, t}}\| \mathbf{z}_{k, t} - 
\pi_\mathbf{K}(\mathbf{p}_k, \boldsymbol{\xi}_t) \|_2
\label{eq:sfm_static}
\end{equation}
where $\mathbf{z}_{k, t}$ is the $k$-th 3D point's corresponding pixel track's 2D coordinates at time $t$, 
$\mathbf{w_{k, t}} \in \{ 0, 1 \}$ is a visibility indicator %
 and $\pi_\mathbf{K}$ is the perspective projection function.

\paragraph{Non-Rigid Bundle Adjustment Term.} 

For dynamic objects, we impose a nonrigid bundle adjustment term, $ E_\mathrm{NR}(\mathcal{P}_\mathrm{dyn}) $, which measures the discrepancy between the dynamic point cloud and pixel tracklets. Here, each pixel tracklet corresponds to a {\it dynamic 3D point sequence}, $ \{\mathbf{p}_{k, t}\}_t $, optimized for each observed tracklet:
\begin{equation}
E_\mathrm{NR}(\mathcal{P}_\mathrm{dyn};\mathcal{C}, \mathcal{Z}, \mathcal{M})
= \sum_{\mathbf{z}_{k} \in \mathcal{M}} \sum_t \mathbf{w_{k, t}}\| \mathbf{z}_{k, t} - 
\pi_\mathbf{K}(\mathbf{p}_{k, t}, \boldsymbol{\xi}_t) \|_2
\end{equation}
where $\mathbf{p}_{k, t} \in \mathbb{R}^3$ is the $k$-th dynamic point's location at $t$.

\begin{figure}[t!]    %
    \centering
    \includegraphics[width=\linewidth]{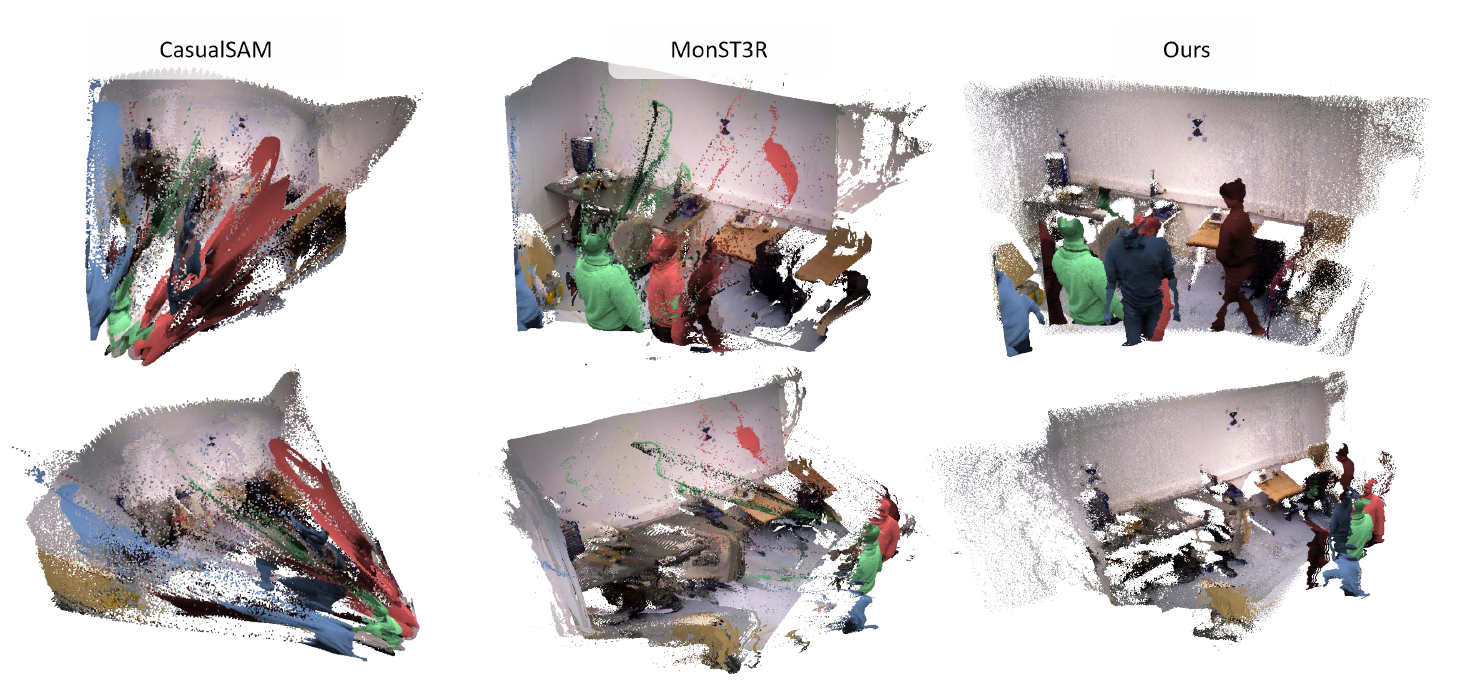}
    \vspace{-5mm}
    \caption{{\bf Qualitative results %
    on Bonn dataset.} Both CasualSAM and MonST3R have trailing artifacts and incorrect dynamic estimations. Uni4D provides clear dynamic and static geometry.  
    }
    \label{fig:bonn}
\end{figure}
\paragraph{Camera Motion Prior.}
Considering that our input is a video, we incorporate a temporal smoothness prior on camera poses that penalizes {sudden changes in relative pose}:
$\boldsymbol{\xi}_{t \rightarrow t+1} = {\boldsymbol{\xi}^{-1}_{t+1}} \cdot \boldsymbol{\xi}_t$. 
We reweight this term based on the magnitude of the relative motion: intuitively, if the relative motion is large, we penalize change rates in relative motion less; if the relative motion is small, we apply a higher penalty on change rate. Formally, we have:
\begin{equation}
    E_\mathrm{cam}(\mathcal{T}) = \sum_t E_\mathrm{rot}(\mathbf R_{t-1, t, t+1}) + \sum_t E_\mathrm{trans}(\mathbf t_{t-1, t, t+1})
\end{equation}
where $E_\mathrm{rot}(\mathbf R_{t-1, t, t+1}) = \frac{2 \|  
\mathrm{rad}(\mathbf{R}_{t\rightarrow t+1}) - 
\mathrm{rad}(\mathbf{R}_{t-1 \rightarrow t}) \|
}{ \| \mathrm{rad}(\mathbf{R}_{t-1 \rightarrow t}) \| + \| \mathrm{rad}(\mathbf{R}_{t\rightarrow t+1}) \|}$ and 
$E_\mathrm{trans}(\mathbf t_{t-1, t, t+1}) = 
\frac{2 \|  
\mathbf{t}_{t\rightarrow t+1} - 
\mathbf{t}_{t-1 \rightarrow t} \|
}{ \| \mathbf{t}_{t-1 \rightarrow t} \| + \| \mathbf{t}_{t\rightarrow t+1} \|}$; $\mathrm{rad}$ converts the rotation matrix into absolute radians.

\paragraph{Dynamic Motion Prior.}

$E_\mathrm{motion}(\mathcal{P}_\mathrm{dyn})$ is a regularization term that encodes the characteristics of the dynamic structure. It contains two prior terms that are used to regularize the dynamic structure, both of which have demonstrated effectiveness in previous work ~\cite{newcombe2015dynamicfusion, sorkine2007rigid}: 
\begin{equation}
    E_\mathrm{motion}(\mathcal{P}_\mathrm{dyn}) = E_\mathrm{arap}(\mathcal{P}_\mathrm{dyn}) + E_\mathrm{smooth}(\mathcal{P}_\mathrm{dyn}).
\end{equation}
$E_\mathrm{arap}$ is an as-rigid-as-possible (ARAP)~\cite{sorkine2007rigid} prior that penalizes extreme deformations that compromise local rigidity. Specifically, we obtain the nearest neighbors of each dynamic control point $k$ by applying KNN over the other tracks and enforce that the relative distances between these close-by pairs do not undergo sudden changes: 
\begin{equation}
E_\mathrm{arap} =  \sum\limits_{\mathbf{t}} \sum\limits_{(k, m)} \mathbf{w_{km,t}} \| d(\mathbf{\mathbf{p}_{k,t}}, \mathbf{p}_{m,t}) - d(\mathbf{\mathbf{p}_{k,t+1}}, \mathbf{p}_{m,t+1}) \|_2
\end{equation}
where $d(\cdot, \cdot)$ is the L2 distance and $\mathbf{w_{km, t}} = 1$ if all relevant points are visible.  

$E_\mathrm{smooth}$ is a simple smoothness term that promotes temporal smoothness for the dynamic point cloud:
\begin{equation}
E_\mathrm{smooth} = \sum\limits_{\mathbf{t}} \sum\limits_{\mathbf{p}_k \in \mathcal{P}_\mathrm{dyn}} \mathbf{w_{k, t}}\|\mathbf{p}_{k,t} - \mathbf{p}_{k,t+1} \|_2. 
\end{equation}

Despite simplicity, both motion terms are crucial in our formulation, as they significantly reduce ambiguities in 4D dynamic structure estimation, which is highly ill-posed. 
Unlike other methods, we do not assume strong model-based motion priors, such as rigid motion~\cite{ma2019deep}, articulated motion~\cite{yang2022banmo}, or a linear motion basis~\cite{wang2024shape}.

\begin{figure*}[t]    %
    \centering
    \vspace{-5mm}
    \includegraphics[width=.9\textwidth]{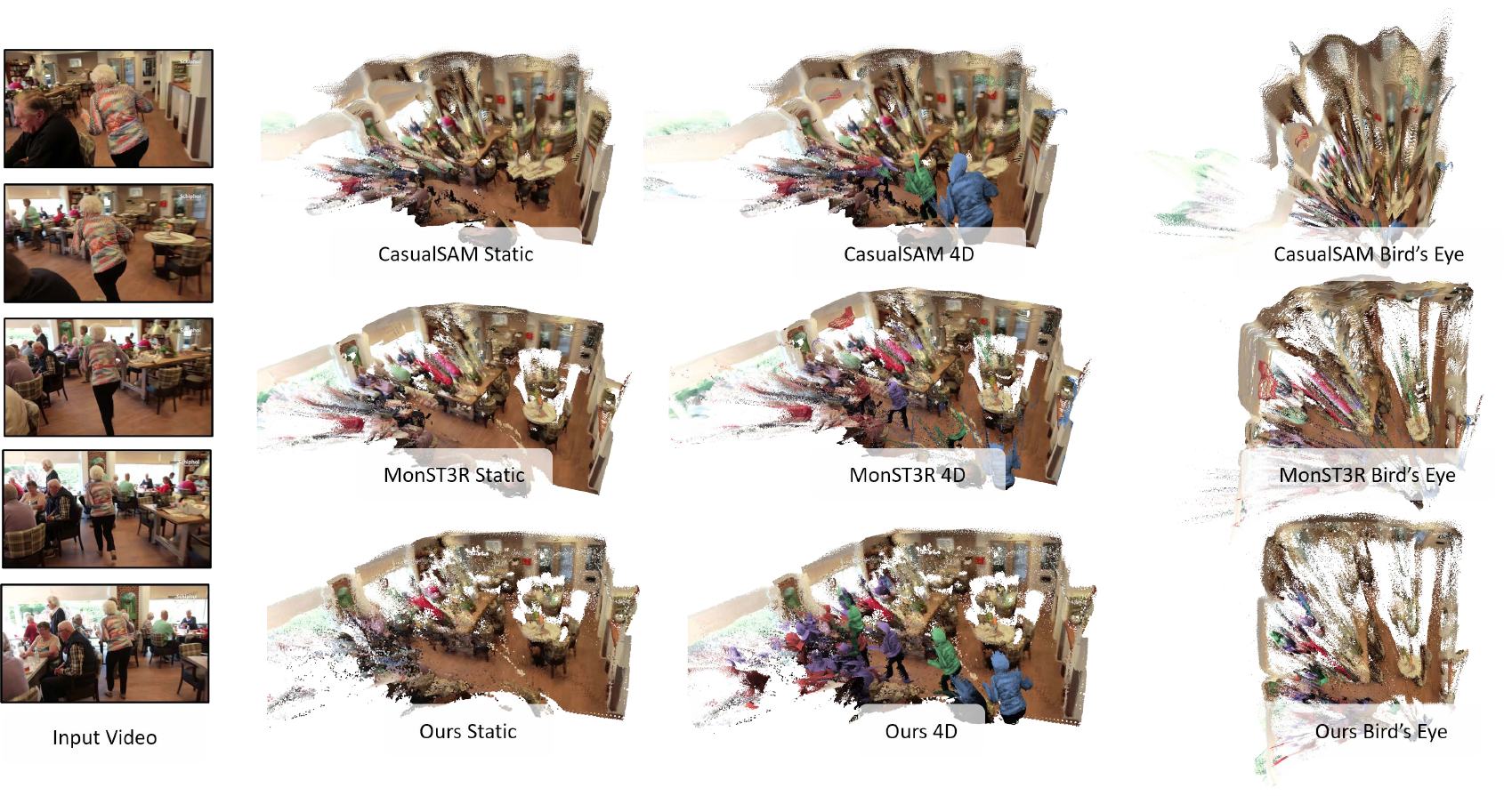}
    \vspace{-5mm}
    \caption{{\bf Qualitative results of 4D reconstruction on DAVIS dataset.} CasualSAM~\cite{zhang2022structure} distorts the room geometry as evident from the bird's eye view. Dynamic elements have inconsistent geometry over time. MonST3R~\cite{zhang2024monst3r} has noisy static geometry in the room's far corner and incomplete dynamic object geometry evident in the green highlight. Uni4D has the cleanest dynamic object reconstruction and segmentation results, with geometrically accurate room shapes as evident from the bird's eye view.}
    \label{fig:qual_davis_lady}
    \vspace{-5mm}
\end{figure*}

\subsection{Inference}\label{sec:inference}

Directly minimizing the energy defined in Eq.~\ref{eq:total} is non-trivial, as our energy function is highly non-linear and involves millions of free variables. To address this, we developed a three-stage optimization pipeline, enabling us to minimize the energy and estimate the scene variables in a divide-and-conquer fashion.

\paragraph{Stage 1: Camera Initialization.}
\label{para:initialization}We start by initializing camera parameters. Combining video depth estimation $\mathcal{D}$ and dense pixel motion $\mathcal{Z}$ allows us to establish 2D-to-3D correspondences. This allows us to initialize and tune $\mathcal{C}$ by minimizing the following energy function with respect to camera parameters {\it only}. Specifically, we can unproject each video frame's depth at time $t$ back to 3D and minimize the following energy function:
\begin{equation}
    \label{eq:camera_init}
    \min_{\mathbf{C}}
    \sum_{(t^\prime, t)} 
    \sum_{\mathbf{z}_k \in \neg\mathcal{M}}
    \| \mathbf{z}_{k, t^\prime} - \pi_\mathbf{K}(\pi^{-1}_\mathbf{K}(\mathbf{z}_{k,t},\mathbf{D}_{t},\boldsymbol{\xi}_{t}), \boldsymbol{\xi}_{t^\prime}) \|_2^2
\end{equation}
where $\pi^{-1}_\mathbf{K}$ is the unprojection function that maps 2D coordinates into 3D world coordinates using estimated depth $\mathbf{D}_{t}$. 
We perform this over all pairs within a temporal sliding window of 5 frames, producing a good initial pose estimate as shown in Tab.~\ref{tab:ablate_multi}.
Given camera initialization $\hat{\mathcal{C}}$, we directly unproject our depth prediction into a common world coordinate system, which provides an initial 4D structure $\hat{\mathcal{P}}$. This is used as initialization for later optimization. 

\paragraph{Stage 2: Bundle Adjustment.}
\label{para:static_bundler}
Our second stage jointly optimizes camera pose and static geometry by minimizing the static component-related energy in a bundle adjustment fashion. Formally speaking, we solve the following: 
\begin{equation}
    \label{eq:static_energy}
    \min_{\mathcal{C}, \mathbf{P}_\mathrm{static}}  E_\mathrm{BA}(\mathcal{C}, \mathbf{P}_\mathrm{static}; \mathcal{Z}, \mathcal{M}) + E_\mathrm{cam}(\mathcal{T})
\end{equation}
By enforcing consistency with each other, this improves both the static geometry and the camera pose quality.
We perform a final scene integration by unprojecting correspondences into 3D using improved pose and filtering outlier noisy points in 3D.
{

}

\paragraph{Stage 3: Non-Rigid Bundle Adjustment.}
\label{para:dyn}
Given the estimated camera pose, our third stage focuses on inferring dynamic structure. Note that we freeze camera parameters in this stage, as we find that incorrect geometry and motion evidence often harm camera pose estimation rather than improve it. Additionally, enabling camera pose optimization introduces extra flexibility in this ill-posed problem, harming robustness. Formally speaking, we solve the following: 
\begin{equation}
    \label{eq:dynamic_energy}
    \min_{\mathcal{P}_\mathrm{dyn}}  E_\mathrm{NR}(\mathcal{P}_\mathrm{dyn};\mathbf{C}, \mathcal{Z}, \mathcal{M})
    + E_\mathrm{motion}(\mathcal{P}_\mathrm{dyn})
\end{equation}
{We initialize $\mathcal{P}_\mathrm{dyn}$ using video depth and our optimized camera pose from Stage 2. We scale $E_\mathrm{smooth}$ and $E_\mathrm{arap}$ with constants 10 and 100 respectively which we found empirically led to better dynamics.
}
This energy optimization might still leave some high-energy noisy points, often from incorrect cues, motion boundaries, or occlusions. We filter these outliers based on their energy values in a final step.

\begin{figure}[t]    %
    \centering
    \includegraphics[width=\linewidth]{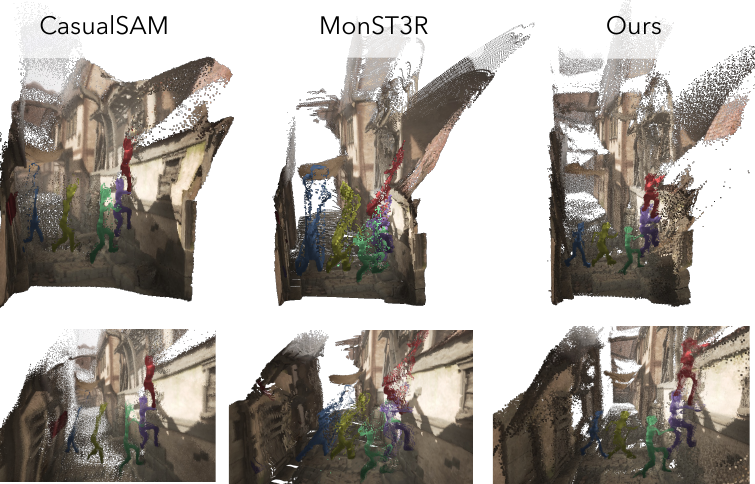}
       \vspace{-5mm}
 \caption{{\bf Qualitative results of 4D reconstruction on Sintel dataset.} We highlight and contrast performance on (1) dynamic objects and (2) geometry on planar surfaces.
    }
    \vspace{-5mm}
    \label{fig:qual_sintel}
\end{figure}

\paragraph{Fusion.}
From our energy minimization, we acquire a semi-dense dynamic and static point cloud along with camera parameters. To further densify the point cloud, enabling each pixel to correspond to a 3D point, we perform depth-based interpolation by computing a scale offset. Details can be found in the supplementary.

To unify our outputs into a consistent 4D representation, we use our camera parameters and static masks to project the aligned depth maps into world coordinates. To handle noisy depth values at boundaries, we create an edge mask filter by thresholding gradients of the depth maps.

This process enables us to update the depth value for each high-confidence pixel, which we can then reproject into 3D space to obtain the final dense dynamic geometry reconstruction result, as shown in Fig.~\ref{fig:teaser} and Fig.~\ref{fig:method}.

\begin{table*}[t]
\centering
\footnotesize
\renewcommand{\arraystretch}{1.3}
\renewcommand{\tabcolsep}{2.5pt}
{
\begin{tabular}{@{}clccc|ccc|ccc@{}}
\toprule
& & \multicolumn{3}{c}{Sintel} & \multicolumn{3}{c}{TUM-dynamics} & \multicolumn{3}{c}{Bonn} \\ 
\cmidrule(lr){3-5} \cmidrule(lr){6-8} \cmidrule(lr){9-11}
{Category} & {Method} & {ATE $\downarrow$} & {RPE trans $\downarrow$} & {RPE rot $\downarrow$} & {ATE $\downarrow$} & {RPE trans $\downarrow$} & {RPE rot $\downarrow$} & {ATE $\downarrow$} & {RPE trans $\downarrow$} & {RPE rot $\downarrow$} \\ 
\midrule
\multirow{2}{*}{{Pose only}} 
& DPVO$^*$\cite{teed2024deep} & 0.171 & 0.063 & 1.291 & 0.019 & 0.014 & 0.406 & 0.022 & 0.014 & 0.913 \\  
& LEAP-VO$^*$\cite{chen2024leap} & \textbf{0.035} & 0.065 & 1.669 & 0.025 & 0.031 & 2.843 & 0.037 & 0.014 & 0.844 \\ 
\midrule 
{{Joint depth}} & Robust-CVD\cite{kopf2021robust} & 0.368 & 0.153 & 3.462 & 0.096 & 0.027 & 2.590 & 0.085 & 0.018 & \textbf{0.803} \\ 
{ \& pose}& CasualSAM\cite{zhang2022structure} & 0.137 & 0.039 & 0.630 & 0.036 & 0.018 & 0.745 & 0.024 & 0.014 & 0.849 \\ 
& Monst3R\cite{zhang2024monst3r} & 0.108 & 0.043 & 0.729 & 0.108 & 0.022 & 1.371 & 0.023 & 0.011 & \underline{0.807} \\ 
& \textbf{Uni4D} & 0.110 & \textbf{0.032} & \underline{0.338} & \textbf{0.012} & \textbf{0.004} & \underline{0.335} & \underline{0.017} & \textbf{0.010} & 0.818 \\
& \textbf{Uni4D*} & \underline{0.092} & \underline{0.033} & \textbf{0.141} & \textbf{0.012} & \textbf{0.004} & \textbf{0.331} & \textbf{0.016} & \textbf{0.010} & 0.817 \\
\bottomrule
\addlinespace[2pt]
\end{tabular}
}
\vspace{-3mm}
\caption{\textbf{Camera Pose Evaluation} on Sintel, TUM-dynamic, and Bonn datasets. We \textbf{bold} and \underline{underline} the best and second best results respectively. $^*$ indicates known camera intrinsic.
}
\label{tab:camera_pose_methods}
\end{table*}

\section{Experiments}
\label{sec:experiments}
\begin{table*}[t]
\centering
\footnotesize
\renewcommand{\arraystretch}{1.2}
\renewcommand{\tabcolsep}{1.5pt}

{
\begin{tabular}{@{}cclcc|cc|cc@{}}
\toprule
 &  &  & \multicolumn{2}{c}{Sintel} & \multicolumn{2}{c}{Bonn} & \multicolumn{2}{c}{KITTI} \\ 
\cmidrule(lr){4-5} \cmidrule(lr){6-7} \cmidrule(lr){8-9}
Alignment & Category & Method & {Abs Rel $\downarrow$} & {$\delta$\textless $1.25\uparrow$} & {Abs Rel $\downarrow$} & {$\delta$\textless $1.25\uparrow$} & {Abs Rel $\downarrow$} & {$\delta$ \textless $1.25\uparrow$} \\ 
\midrule
 \multirow{7}{*}{\begin{minipage}{2.0cm}\centering Per-sequence scale \& shift\end{minipage}} &\multirow{3}{*}{\begin{minipage}{2.0cm}\centering Single-frame depth \end{minipage}} & Metric3D\cite{hu2024metric3d} & \underline{0.205} & 71.9 & 0.044 & \underline{98.5} & \underline{0.039} & \textbf{98.8} \\ 
 && Depth-pro\cite{bochkovskii2024depth} & 0.280 & 60.5 & 0.049 & \textbf{98.6} & 0.080 & 94.2 \\ 
 && Unidepth\cite{piccinelli2024unidepth} & \textbf{0.198} & \textbf{72.8} & \underline{0.040} & \underline{98.5} & \textbf{0.038} & \textbf{98.8} \\ 

\cmidrule{2-9}

 &\multirow{1}{*}{\begin{minipage}{2.0cm}\centering Video depth \end{minipage}}
 
 & DepthCrafter\cite{hu2024depthcrafter} & 0.231 & 69.0 & 0.065 & 97.6 & 0.112 & 88.4 \\ 

\cmidrule{2-9}

 &\multirow{4}{*}{\begin{minipage}{2.0cm}\centering Joint video depth \& pose \end{minipage}}& Robust-CVD\cite{kopf2021robust} & 0.358 & 49.7 & 0.108 & 89.8 & 0.182 & 72.9 \\  && CasualSAM\cite{zhang2022structure} & 0.292 & 56.9 & 0.069 & 96.6 & 0.113 & 88.3 \\ 
 && Monst3r\cite{zhang2024monst3r} & 0.358 & 52.1 & 0.060 & 95.0 & 0.085 & 91.9 \\ 
& & \textbf{Uni4D} & 0.216 & \underline{72.5} & \textbf{0.038} & 98.3 & 0.098 & 89.7 \\ 
\addlinespace[1.5pt]
\hline\hline
\addlinespace[1.5pt]
 \multirow{2}{*}{\begin{minipage}{2.0cm}\centering Per-sequence scale\end{minipage}}
 & Joint depth \& pose & Monst3r\cite{zhang2024monst3r} & 0.344 &55.9 & 0.041 & 98.2 & 0.089 & 91.4 \\ 
 & Joint depth \& pose & \textbf{Uni4D} & \textbf{0.289} & \textbf{64.9} & \textbf{0.038} & \textbf{98.3} & \textbf{0.086} & \textbf{93.3} \\ 
\bottomrule
\end{tabular}
}
\vspace{-3mm}
\caption{\textbf{Video depth evaluation} on Sintel, Bonn, and KITTI datasets. We \textbf{bold} and \underline{underline} the best and second best results respectively.}
\vspace{-5mm}
\label{tab:videodepth_methods}
\end{table*}

Uni4D estimates camera pose, depth, and 3D motion from a single video. We perform experiments evaluating its performance with respect to baselines on these tasks.

\subsection{Implementation Details}

For optimization, we use the Adam optimizer with ReduceLROnPlateau and EarlyStopping in PyTorch~\cite{kingma2014adam, paszke2019pytorch}. We perform 600 iterations per sliding window in stage 1, 2000 iterations in stage 2, and 1000 iterations in stage 3. We initialize the learning rate at \num{1e-3} for stage 1, and \num{1e-2} for stages 2 and 3, reducing all to \num{1e-4}. 
Our entire framework takes roughly 5 minutes for a 50-frame video on a RTX A6000 GPU. We include a detailed runtime breakdown in the supplementary.
We run all baselines on the datasets using their official implementations and hyperparameters.
Co-Trackers~\cite{karaev2024cotracker3} are initialized at a dense 50x50 grid, with a 75x75 grid for Sintel to handle its large camera perspective change. All optimization hyperparameters are kept the same for all runs on all datasets.

\subsection{Pose Estimation}

\paragraph{Baselines.} We compare with several recent methods for pose estimation in dynamic scenes. LEAPVO~\cite{chen2024leap} and DPVO~\cite{teed2024deep} are learning-based visual odometry methods. Robust-CVD~\cite{kopf2021robust} optimizes for pose and depth deformation through an SFM pipeline. CasualSAM~\cite{zhang2022structure} further improves on the idea by directly finetuning network weights along with a novel uncertainty formulation. Monst3R~\cite{zhang2024monst3r} is a very recent work that fine-tunes DUSt3R~\cite{wang2024dust3r} for 4D reconstruction through PnP~\cite{lepetit2009ep}. 

\paragraph{Benchmarks and metrics.} We evaluate pose estimation on three dynamic datasets: Sintel~\cite{butler2012naturalistic}, TUM-dynamics~\cite{sturm2012benchmark}, and Bonn~\cite{palazzolo2019refusion}. We follow LEAP-VO's evaluation split for Sintel and use all videos from TUM-dynamics and Bonn. Following MonST3R~\cite{zhang2024monst3r}, we subsample every 3 frames from the first 270 frames from TUM-dynamics to save compute. We follow the standard pose evaluation process of aligning camera poses with Umeyama alignment~\cite{umeyama1991least}. We report Absolute Translation Error (ATE), Relative Translation and Rotation Error (RPE trans and RPE rot).

\paragraph{Results.} As reported in Tab.~\ref{tab:camera_pose_methods}, Uni4D achieves competitive results across all metrics and datasets, highlighting the generalizability and performance of our pipeline. Uni4D is flexible to the availability of camera intrinsics, showing further improvement with known camera intrinsics. Training-based approaches such as LEAP-VO achieves good results on synthetic dataset like Sintel but does not generalize well to real-world datasets. Our method matches performance with MonST3R~\cite{zhang2024monst3r} on Sintel and achieves significantly better results on real-world datasets even compared to methods using ground-truth intrinsics.

\subsection{Video Depth Evaluation}

\paragraph{Baselines.} For video depth accuracy evaluations, we focus on the top performing metric depth estimators, namely Metric3Dv2~\cite{hu2024metric3d}, Depth-Pro~\cite{yang2024depth}, DepthCrafter~\cite{hu2024depthcrafter} and Unidepth~\cite{piccinelli2024unidepth}. 
Metric depth estimators are trained without scale and shift alignment, making them strong baselines for video depth estimation. %
We also include the same baselines as our pose evaluations for joint 4D modeling approaches, namely CasualSAM~\cite{zhang2022structure}, RCVD~\cite{kopf2021robust}, and MonST3R~\cite{zhang2024monst3r}.

\paragraph{Benchmarks.} We evaluate video depth estimates on Sintel~\cite{butler2012naturalistic}, Bonn~\cite{palazzolo2019refusion} and KITTI~\cite{geiger2013vision}. We follow standard video depth evaluation protocols~\cite{hu2024depthcrafter} of aligning global shift and scale to predicted video depthmaps. We report the absolute relative error (Abs Rel) and percent of inlier points ($\delta < 1.25$). All methods are aligned in disparity space using the same least-squares alignment. We additionally report scale-aligned depth estimates similar to MonST3R.

\paragraph{Results.} By leveraging the strong performance of single-frame metric depth estimation models, Tab.~\ref{tab:videodepth_methods} shows Uni4D achieves competitive depth estimation results, with superior performance among joint depth and pose methods, and closely matching the performance of single-frame depth estimation models on some datasets. With per-sequence scale, our model produces more accurate depth estimates across all datasets compared to recently released MonST3R.
Our method closely retains depth estimation accuracy of underlying Unidepth depthmaps, while significantly improving its depth consistency shown in Sec.~\ref{subsec:ablation}.

\subsection{Qualitative}

We further show qualitative results of our reconstructions on the DAVIS~\cite{perazzi2016benchmark} dataset. We apply MonST3R's own confidence-guided fusion for their reconstruction. We fuse casualSAM's depthmaps with dynamic masking by thresholding its uncertainty prediction to output 4D reconstruction. We use highlights to indicate dynamic objects at different timesteps. Throughout our qualitative results in Figs.~\ref{fig:qual_davis_bear},\ref{fig:bonn},\ref{fig:qual_davis_lady},\ref{fig:qual_sintel}, CasualSAM produces warped geometry and poor dynamic segmentations. MonST3R produces poor geometry in far regions and noisy dynamic masks and shapes. Our model produces the cleanest dynamic segmentations, along with the best dynamic and static geometry.

\begin{figure}[t!]    %
    \centering
    \includegraphics[width=\linewidth]{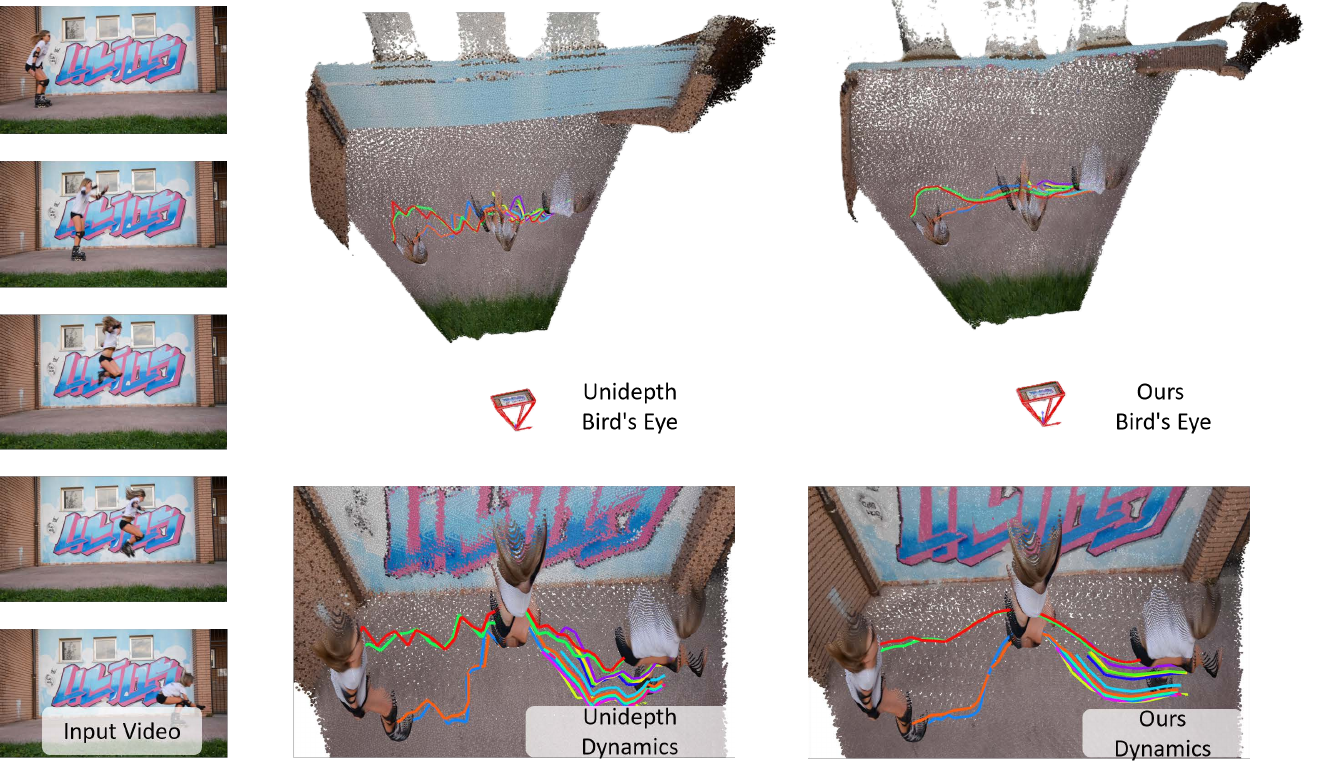}
        \vspace{-7mm}
    \caption{{\bf Depth Consistency.} Direct fusion of Unidepth~\cite{piccinelli2024unidepth} predictions causes misaligned scene geometry visible as a layered wall in bird's-eye view, and jittering dynamic motion. Through motion priors and alignment, Uni4D produces a thin, crisp wall structure and smooth dynamic motion.}
    \label{fig:ablate_cons}
    \vspace{-3mm}
\end{figure}

\subsection{Ablation Study}
\label{subsec:ablation}

We ablate our {\bf multi-stage optimization} in Tab.~\ref{tab:ablate_multi}, highlighting the importance of each stage. Joint 4D optimization adds complexity that requires strong initialization for optimal convergence. Stage 1 introduces drift, which Stage 2 rectifies, resulting in superior final pose estimates.

Directly reprojecting Unidepth depth maps leads to flickering geometry. We ablate our fused depthmap quantitatively in Tab.~\ref{tab:ablate_consistency}, showing that Uni4D significantly improves consistency using Self-Consistency (SC) metrics~\cite{khan2023temporally}. SC measures depth errors between estimated and reprojected depth maps in static regions.
Furthermore, we qualitatively show the effectiveness of Uni4D in rectifying Unidepth inconsistencies in Fig.~\ref{fig:ablate_cons}. Our reconstruction achieves much better geometric and temporal consistency over Unidepth. Please refer to the supplementary for more ablative results regarding choice of foundation models.

\begin{table}[t]
\centering
\footnotesize
\begin{tabular}{@{}lccc}
\toprule
 & \multicolumn{3}{c}{Sintel} \\ 
\cmidrule(lr){2-4}
Method & {SC $\downarrow$} & {$\delta_{SC}$\textless $0.01\uparrow$} & {$\delta_{SC}$\textless $0.05\uparrow$} \\ 
\midrule
Unidepth~\cite{piccinelli2024unidepth} & 0.109 & 31.8 & 76.8 \\ 
\textbf{Uni4D} & 0.043 & 69.3 & 88.1 \\ 
\bottomrule
\end{tabular}
\vspace{-3mm}
\caption{\textbf{Video depth consistency} on Sintel. Uni4D improves Unidepth in consistency. 
}
\label{tab:ablate_consistency}
\vspace{-3mm}
\end{table}

\begin{table}[t]
\centering
\footnotesize
\begin{tabular}{@{}lccc@{}}
\toprule
 & \multicolumn{3}{c}{Sintel} \\ 
\cmidrule(lr){2-4}
Method &  {ATE $\downarrow$} & {RPE trans $\downarrow$} & {RPE rot $\downarrow$} \\ 
\midrule
Uni4D (stage 1 only) & 0.150 & 0.051 & 0.551 \\ 
Uni4D (stage 2 only) & 0.587 & 0.193 & 4.12 \\
\textbf {Uni4D (full)} & \textbf{0.110} & \textbf{0.032} & \textbf{0.338} \\
\bottomrule
\end{tabular}
\vspace{-3mm}
\caption{\textbf{Multi-stage Ablation.} We evaluate pose estimation results on Sintel for both stage 1 and stage 2.}
\label{tab:ablate_multi}
\vspace{-3mm}
\end{table}

\section{Conclusion}
\label{sec:conclusion}

This paper presents Uni4D, a framework unifying visual foundation models and structured energy minimization for dynamic 4D modeling from casual video. Our key insight is to optimize a 4D representation that aligns with visual cues from foundation models while following motion and geometry priors. Results show state-of-the-art performance with superior visual quality on Sintel, DAVIS, TUM-Dynamics and Bonn datasets, without any retraining or fine-tuning.

\section*{Acknowledgements}
This project is supported by the Intel AI SRS gift, Amazon-Illinois AICE grant, Meta Research Grant, IBM IIDAI Grant, and NSF Awards \#2331878, \#2340254, \#2312102, \#2414227, and \#2404385. We greatly appreciate the NCSA for providing computing resources.

{
    \small
    \bibliographystyle{ieeenat_fullname}
    \bibliography{main}
}

\clearpage
\setcounter{page}{1}
\maketitlesupplementary

\appendix

\section{Additional Qualitative Results}

\textbf{We provide extensive qualitative results of Uni4D and other baselines on all datasets \href{https://uni4d-paper.github.io/uni4d-vis/}{here}}. We ran MonST3R using their provided hyperparameters from their respective official codebase on all datasets. In our qualitative comparison, we use the \textit{estimated} dynamic masks from MonST3R. This ensures a fair comparison, as the qualitative results for ALL competing algorithms, including ours and all the baselines, do not use privileged information. We generate dynamic masks from CasualSAM by thresholding its uncertainty prediction, using their estimated video depth maps and camera pose to output 4D reconstruction. For Uni4D, we use the same set of hyperparameters throughout our pipeline for all videos for each respective dataset. \\

All reconstructions are performed with depth estimates resized back to original input resolutions, and with background point clouds downsampled 5 times for efficiency using uniform downsampling. We render final (point-cloud) reconstructions using Open3D, manually picking similar viewpoints for all methods since the reconstructions are neither axis nor scale aligned. We provide visualizations for DAVIS~\cite{perazzi2016benchmark}, Sintel~\cite{butler2012naturalistic}, TUM-dynamics~\cite{sturm2012benchmark}, Bonn~\cite{palazzolo2019refusion}, and KITTI~\cite{geiger2013vision}, including failure cases. We include sampled frames of our visualizations in Fig.~\ref{fig:davis_supp_qual},~\ref{fig:sintel_tumd_supp_qual},~\ref{fig:bon_kitti_supp_qual}, though we strongly encourage viewing the attached webpage for the best visualization experience of our results.

\section{Quantitative Evaluation Procedures}
{For all quantitative evaluation results of pose and video depth maps, we follow MonST3R's evaluation script.  We ran all of our baselines using their official codebase and default hyperparameters on all datasets. We use the same depth map alignment, based on least squares in disparity space, for all our depth map evaluations. This is slightly different from the evaluation in MonST3R, where after confirming with MonST3R author, different alignment methods were used for different baselines. This accounts for the different quantitative results in our study and MonST3R's for overlapping baselines (Particularly, we found that CasualSAM~\cite{zhang2022structure} and DepthCrafter~\cite{hu2024depthcrafter} achieves better reported performance than in the MonST3R paper (see Table 2 in main paper)). }

\section{Runtime Breakdown}

\vspace{-2mm}
\begin{figure}[htbp]    %
    \centering
    \includegraphics[width=\columnwidth]{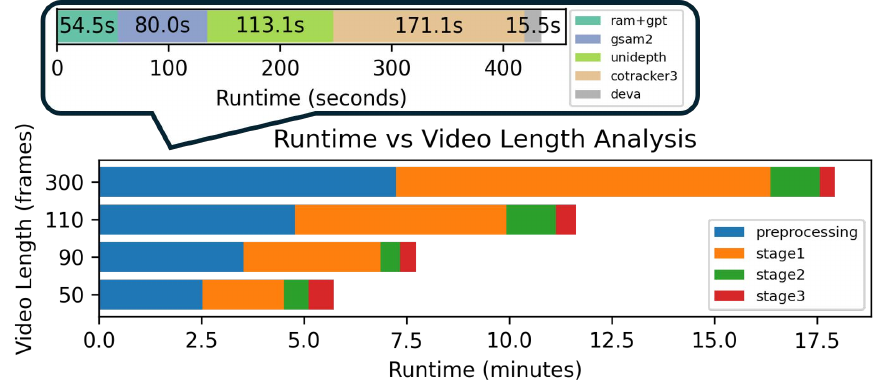}
    \vspace{-5mm}
    \caption{\textbf{Runtime Breakdown of preprocessing and optimization}}
    \label{fig:runtime}
    \vspace{-5mm}
\end{figure}

Figure~\ref{fig:runtime} presents a detailed runtime breakdown of Uni4D's preprocessing and optimization stages. Runtimes are averaged across videos of the same length from our evaluation datasets. The reliance on foundation models significantly contributes to the preprocessing time, particularly due to Unidepth~\cite{piccinelli2025unidepthv2} and CotrackerV3~\cite{karaev2024cotracker3}. Stage 1 initialization, which estimates poses from scratch, accounts for the majority of the optimization runtime. Overall, runtime scales linearly. Further improvements through advanced optimizers and parallelization are left for future work.

\section{Densification Details}

During fusion, we wish to densify the sparse depth obtained from our point trajectories to obtain full-resolution depth maps. Naively interpolating our projected depth in image space leads to poor results, especially across edges and boundaries. Fortunately, flickers in predicted depth maps are usually constant across each scene element. Using this observation, we perform a scale interpolation derived in 3D to obtain a scaling correction $s(\mathbf{x})$ for pixel coordinates $\mathbf{x}$ for every pixel in the depth maps using the following interpolation formula:

\begin{equation}
s(\mathbf{x}) = \sum_{\mathbf{p_i} \in \mathbf{n(x)}} \mathbf{w_i} \frac{z(\mathbf{p_i}, \mathbf{\xi_t})}{\mathbf{D_t(\pi_\mathbf{K}(\mathbf{p_i}, \mathbf{\xi_t}))}} 
\end{equation}

where $\mathbf{n(x)}$ are the 3 nearest point trajectories in 3D of the unprojection of $\mathbf{x}$, $\pi_\mathbf{K}^{-1}(\mathbf{x},\xi_t)$. $\mathbf{w_i}$ is simply $\frac{1}{d_i}$ where $d_i$ is the euclidean distance between unprojection of $\mathbf{x}$ and each corresponding $\mathbf{p_i}$. $z(\mathbf{p_i}, \mathbf{\xi_t})$ returns the z-component of $\mathbf{p_i}$ after transforming to  camera coordinates at time $\mathbf{t}$, and $\mathbf{D_t()}$ returns the depth value from our estimated video depth at the given pixel coordinate at time $\mathbf{t}$. We get our final depth value at pixel $\mathbf{x}$ through $s(\mathbf{x}) \cdot \mathbf{D_t(x)}$. Note that our interpolation is tracklet-aware and searches for nearest neighbors within our preprocessed dynamic object masks. Intuitively, this performs depth map alignment by aligning the original temporally inconsistent depth predictions with our point trajectories to achieve consistent and stable video depth.

\begin{table}
\centering
\footnotesize
\begin{tabular}{@{}lccc@{}}
\toprule
 & \multicolumn{3}{c}{Sintel} \\ 
\cmidrule(lr){2-4}
Method &  {ATE $\downarrow$} & {RPE trans $\downarrow$} & {RPE rot $\downarrow$} \\ 
\midrule
Uni4D (Metric3D~\cite{hu2024metric3d}) & 0.135 & 0.033 & \textbf{0.347} \\ 
Uni4D (Depth-Pro~\cite{bochkovskii2024depth}) & 0.143 & \textbf{0.032} & 0.451 \\
Uni4D (Depthanythingv2-outdoor~\cite{yang2024depth}) & 0.112 & 0.040 & 0.556 \\
\textbf {Uni4D (Unidepth)} & \textbf{0.109} & \textbf{0.032} & \textbf{0.347} \\
\bottomrule
\end{tabular}
\vspace{-3mm}
\caption{\textbf{Performance with different depth models.} We evaluate pose estimation performance on Sintel using different metric depth estimation models.}
\label{tab:ablate_diff_depth}
\vspace{-3mm}
\end{table}

\section{Depth Model Ablation Study}

A key strength of Uni4D is that its modular pipeline allows for the interchangeability of its underlying pre-trained components. We try different depth estimation models and evaluate their pose and depth estimation results on the Sintel~\cite{butler2012naturalistic} dataset in Tab.~\ref{tab:ablate_diff_depth}. We find that currently, Unidepth~\cite{piccinelli2024unidepth} provides the best results.

\section{Ablation on tracker and segmentation choice}

\vspace{-3mm}
\begin{table}[htbp]
\centering
\footnotesize
\begin{tabular}{@{}l@{\hspace{6pt}}*{3}{c@{\hspace{6pt}}}|*{2}{c@{\hspace{6pt}}}@{}}
\toprule
Method & ATE$\downarrow$ & RPE-t$\downarrow$ & RPE-r$\downarrow$ & AbsRel$\downarrow$ & $\delta_{1.25}\uparrow$ \\ 
\midrule
Uni4D (TAPIR) & 0.131 & 0.048 & 1.483 & 0.224 & 71.7 \\
Uni4D (BootsTAPIR) & 0.135 & \textbf{0.027} & 0.403 & 0.219 & \underline{72.5}   \\ 
Uni4D (CTv2) & \underline{0.111} & 0.032 & \textbf{0.309} & \textbf{0.214} & \textbf{72.7}  \\
Uni4D (original, CTv3) & \textbf{0.110} & \underline{0.031} & \underline{0.338} & \underline{0.216} & \underline{72.5} \\
\midrule
Uni4D (Mask-RCNN) & \textbf{0.107} & \bf{0.028} & \underline{0.498} & \underline{0.269} & \underline{68.2} \\
Uni4D (original, DEVA) & \underline{0.110} & \underline{0.031} & {\bf 0.338} & {\bf 0.216} & {\bf 72.5} \\
\bottomrule
\end{tabular}
\caption{\textbf{Ablation on different trackers and segmentators} We compare both pose and geometry performance on Sintel using different tracklet and segmentation models.}
\label{tab:diff_track_seg}
\vspace{-5mm}
\end{table}

We compare different trackers and segmentors in Tab. ~\ref{tab:diff_track_seg}. TAPIR and BootsTAPIR lead to worse camera pose and depth. CTv2 (CotrackerV2) performs similarly to CTv3 (CotrackerV3), though we found CTv3 to have better dynamic correspondences qualitatively. Mask-RCNN tends to have false positives, leading to over filtering of static tracklets. Due to our dense tracklet initialization, this does not necessarily harm pose estimation. However, it harms depth estimation due to our tracklet-aware densification.

\begin{table}
\centering
\footnotesize
\begin{tabular}{@{}lccc@{}}
\toprule
 & \multicolumn{2}{c}{Sintel} \\ 
\cmidrule(lr){2-3}
Method &  {Abs Rel $\downarrow$} & {$\delta$\textless $1.25\uparrow$}\\ 
\midrule
Unidepth~\cite{piccinelli2024unidepth} & \underline{0.178} & 78.4 \\
Uni4D (no dynamic opt.)  & 0.253 & 75.1 \\ 
Uni4D ( $+\mathbf{E_\mathrm{smooth}}$) & 0.228 & 77.0  \\
Uni4D ( $+\mathbf{E_\mathrm{smooth}}+\mathbf{E_\mathrm{arap}}$) & 0.226 & 77.1  \\
Uni4D ($+\mathbf{E_\mathrm{smooth}}+\mathbf{E_\mathrm{arap}}+\mathbf{E_\mathrm{NR}}$) & 0.220 & \underline{78.8}  \\
Uni4D with gt seg ($+\mathbf{E_\mathrm{smooth}}+\mathbf{E_\mathrm{arap}}+\mathbf{E_\mathrm{NR}}$) & \textbf{0.169} & \textbf{79.4}  \\

\bottomrule
\end{tabular}
\vspace{-3mm}
\caption{\textbf{Ablation on $\mathbf{E_\mathrm{motion}}$ ($\mathbf{E_\mathrm{arap}}$, $\mathbf{E_\mathrm{smooth}}$) and $\mathbf{E_\mathrm{NR}}$}. We ablate on our different dynamic element energy terms $\mathbf{E_\mathrm{motion}}$ and $\mathbf{E_\mathrm{NR}}$ through depth map accuracy on Sintel (only considering dynamic elements as defined by ground truth dynamic masks).}
\label{tab:ablate_diff_dyn_energy}
\vspace{-3mm}
\end{table}

\section{Dynamic Regularization Ablation Study}

We ablate our different energy terms for dynamic objects in Tab.~\ref{tab:ablate_diff_dyn_energy}, demonstrating depth map improvements in dynamic regions with each additional dynamic energy term. Note that dynamic segmentations are particularly difficult on Sintel dataset due to large camera motions and close-ups of dynamic elements. Despite the challenging setting, our method produces better dynamic depth maps under the $\delta$\textless $1.25$ metric with estimated dynamic segmentations. With ground truth dynamic masks, our dynamic regularization improves on depth map estimation in dynamic regions over Unidepth~\cite{piccinelli2024unidepth}.

\section{Qualitative Results on Camera Pose Evaluation}

\begin{figure*}[t]    %
    \centering
    \includegraphics[width=\linewidth]{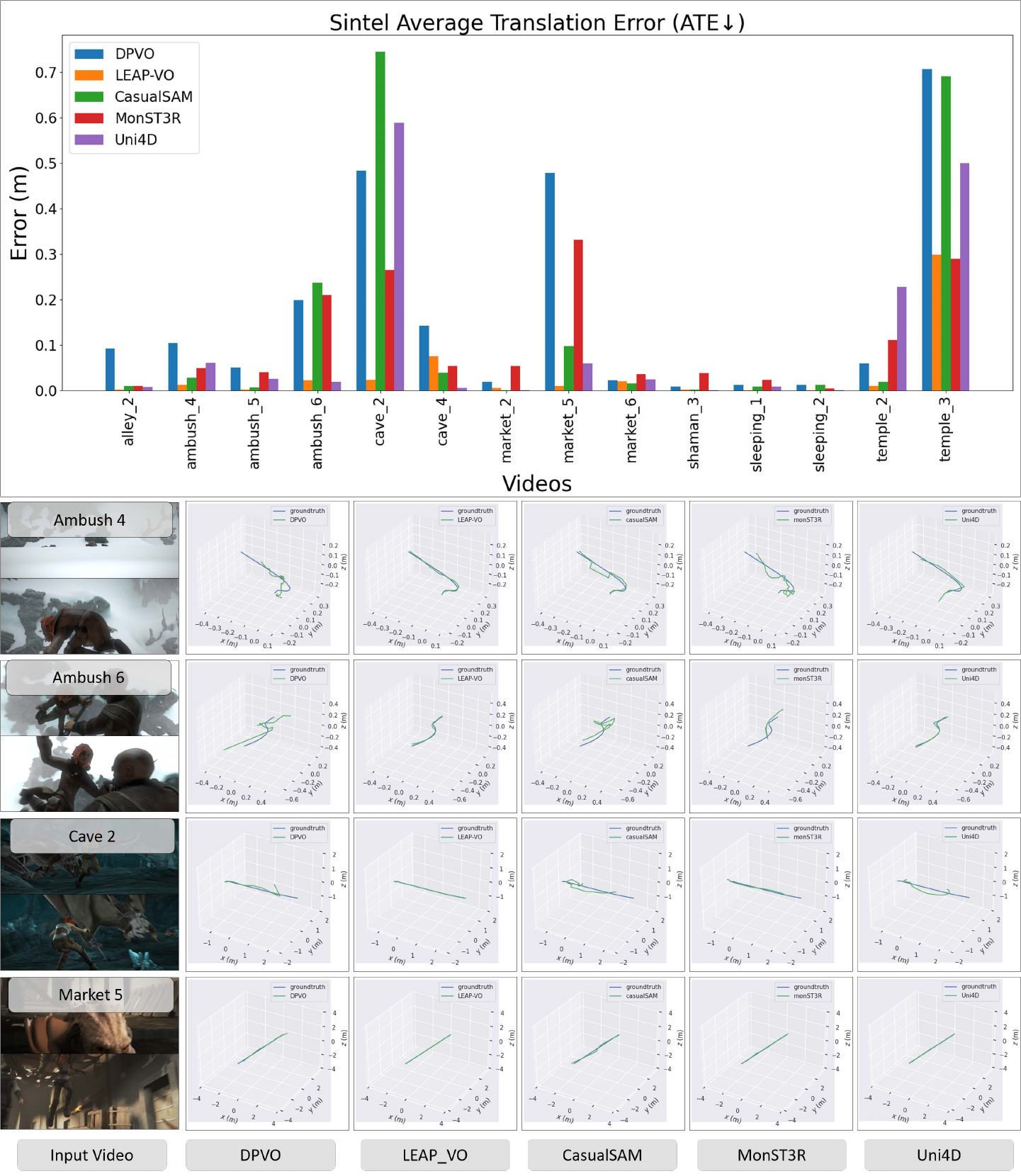}
 \caption{{\bf Qualitative Pose Results on Sintel} Uni4D provides accurate pose estimate on Sintel which contains highly dynamic elements which takes up much of the frame, with 2 failure cases in cave 2 and temple 3. Cave 2 full visualization can be seen from our webpage under "failure cases". Other pose estimates are competitive and even outperform baselines in certain scenes.
    }
    \label{fig:sintel_ATE_qual}
\end{figure*}

\begin{figure*}[t]    %
    \centering
    \includegraphics[width=\linewidth]{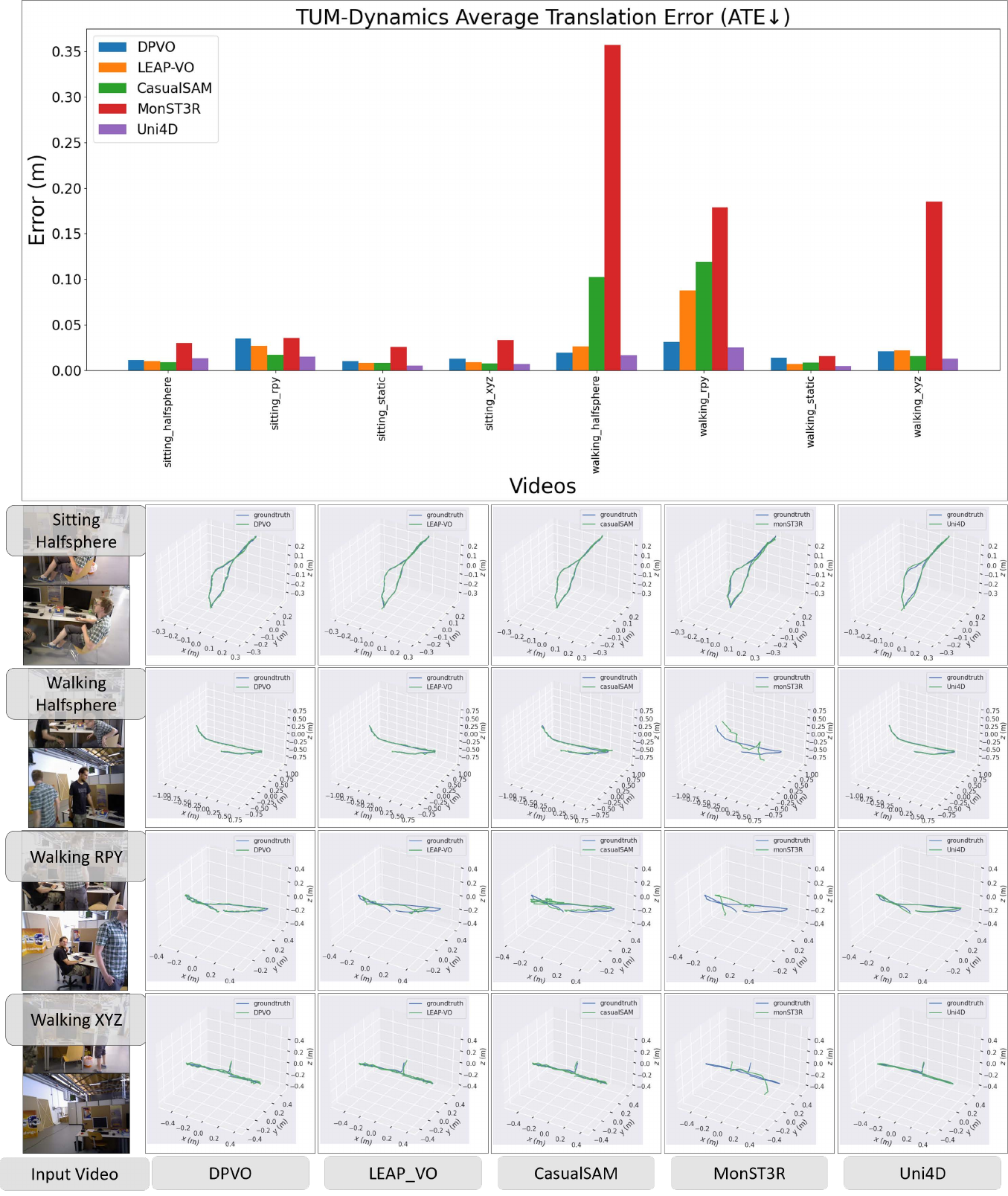}
 \caption{{\bf Qualitative Pose Results on TUM-Dynamics} Uni4D performs well in real-world datasets due to its leverage of big models. Across varied settings where camera motion is mainly rotations (rpy videos), static (static videos), and contains highly dynamic elements (walking videos), Uni4D surpasses other baselines in estimating accurate camera pose.
    }
    \label{fig:tumd_ATE_qual}
\end{figure*}

\begin{figure*}[t]    %
    \centering
    \includegraphics[width=\linewidth]{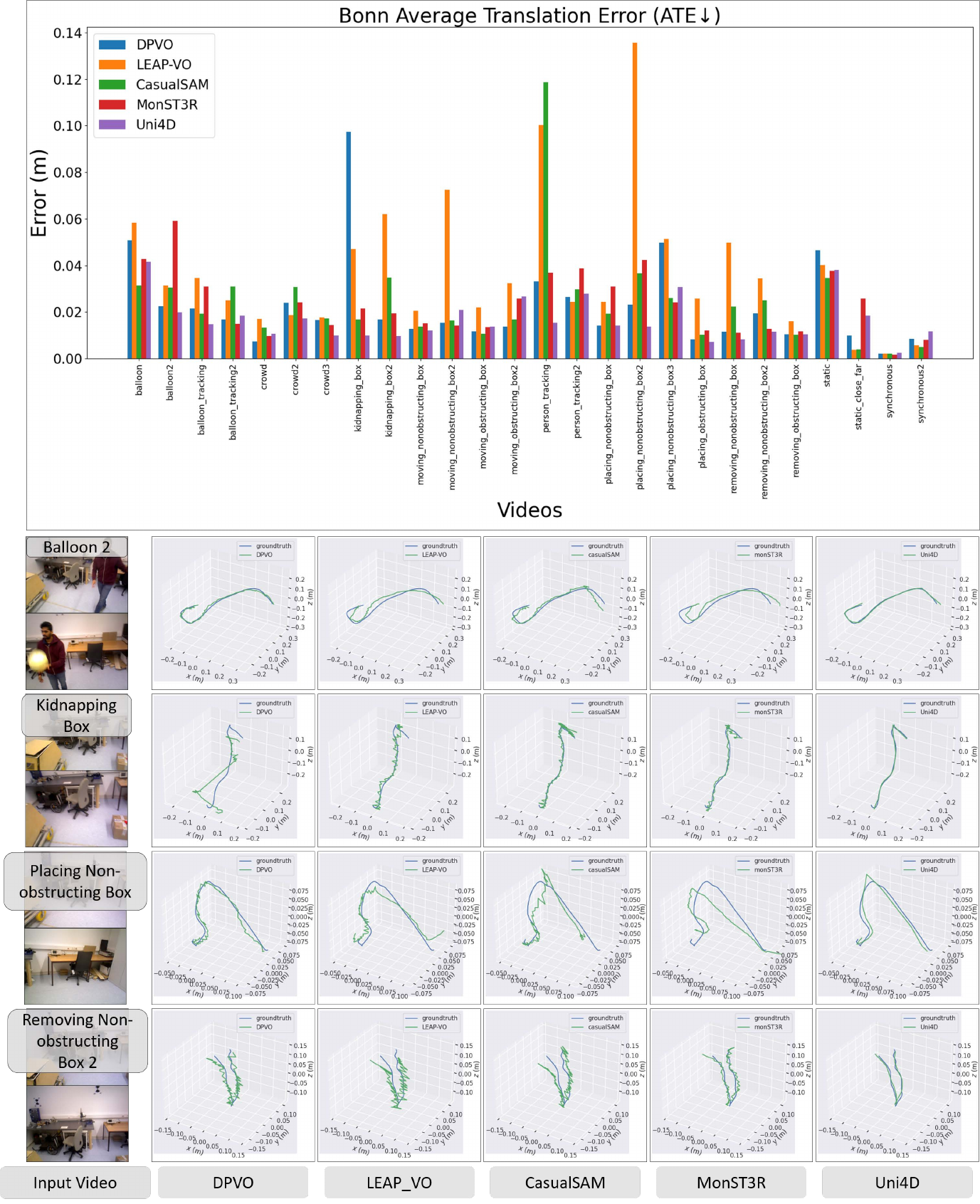}
 \caption{{\bf Qualitative Pose Results on Bonn} Uni4D performs well in real-world datasets, with minimal trajectory errors across all videos in Bonn dataset, successfully estimating trajectories in difficult videos such as 'kidnapping box' and 'placing non-obstructing box' where other baselines face difficulties in.
    }
    \label{fig:bonn_ATE_qual}
\end{figure*}

For a thorough breakdown and visualization of our camera pose evaluations, we plot our Average Translation Error (ATE) results on all camera pose datasets in Fig.~\ref{fig:sintel_ATE_qual} \ref{fig:tumd_ATE_qual} \ref{fig:bonn_ATE_qual}. Despite the highly dynamic nature of the Sintel dataset~\cite{butler2012naturalistic}, Uni4D provides accurate estimations for most videos thanks to accurate dynamic segmentation, with failure cases in Cave 2 and Temple 3 as seen in Fig.\ref{fig:sintel_ATE_qual}. Both videos have large dynamic objects that make them challenging among other baselines as well. For real-world datasets TUM-Dynamics~\cite{sturm2012benchmark} and Bonn~\cite{palazzolo2019refusion}, Uni4D consistently produces the best camera pose estimates with minimal failure cases. Note that across the diverse settings in TUM-Dynamics, including purely translational, rotational, and static camera motion, Uni4D nearly always provides the best pose estimates as seen in (Fig.~\ref{fig:tumd_ATE_qual}). Our camera smoothness regularization also results in the smoothest trajectories, as shown in Fig.~\ref{fig:bonn_ATE_qual}.

\section{Failure Cases}

We provide full visualization of failure cases in our webpage, and sampled frames in Fig.~\ref{fig:failure_cases}. Failure cases include erroneous dynamic masks, depth map estimations, and localization. These errors stem from the underlying models used for segmentation, depth map estimation, and pixel tracking respectively. As the various models are improved upon in the future, we can expect the performance of Uni4D to improve as well.

\begin{table*}[t]
    \centering
    \setlength\tabcolsep{0.05em} %
    \vspace{-2mm}\resizebox{\textwidth}{!}{
    \begin{tabular}{ccc}
        \hline 
          \multicolumn{3}{c}{\textbf{SoapBox}} \\
    \multicolumn{1}{c}{\textbf{\underline{CasualSAM}}} & \multicolumn{1}{c}{\textbf{\underline{MonST3R}}} & \multicolumn{1}{c}{\textbf{\underline{Uni4D (Ours)}}} \\
          
          \includegraphics[width=0.33\textwidth, clip]{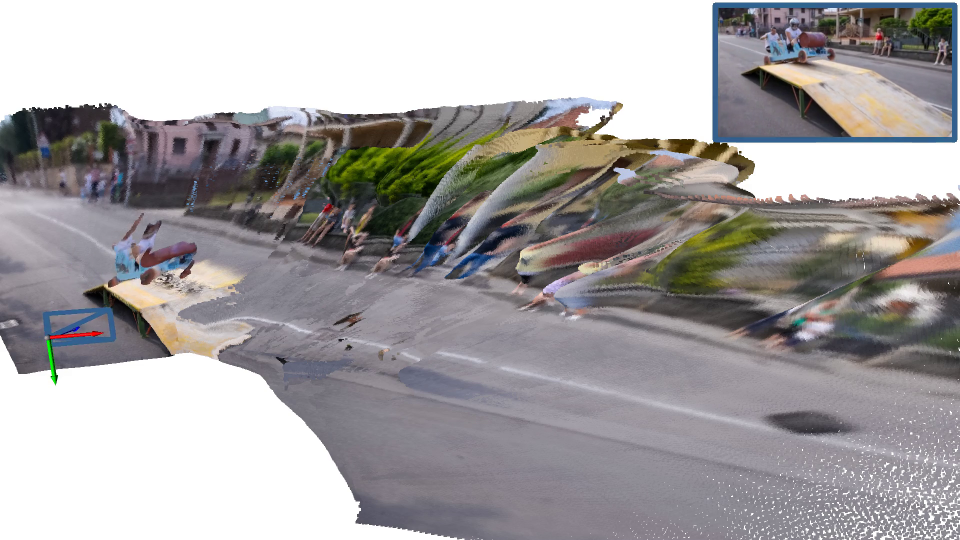} & 
          \includegraphics[width=0.33\textwidth, clip]{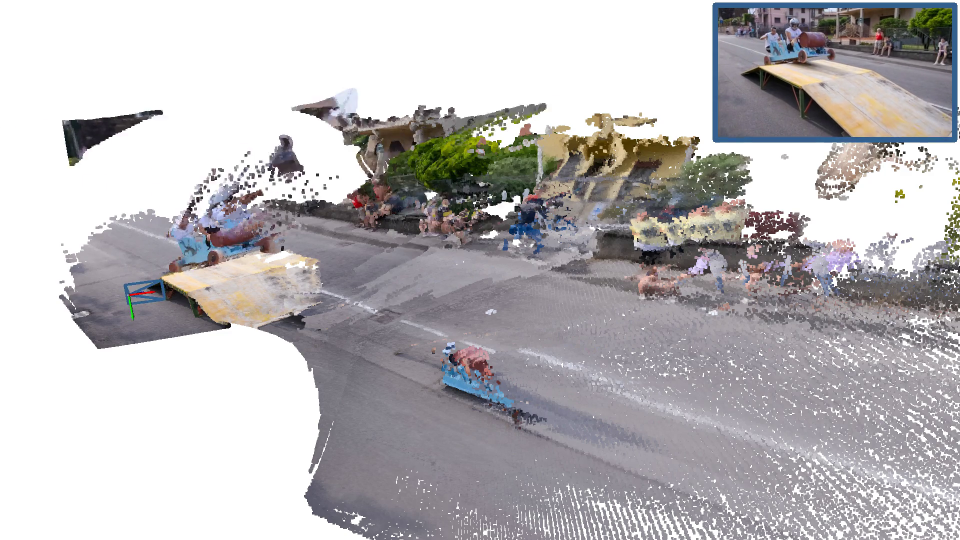} &
          \includegraphics[width=0.33\textwidth, clip]{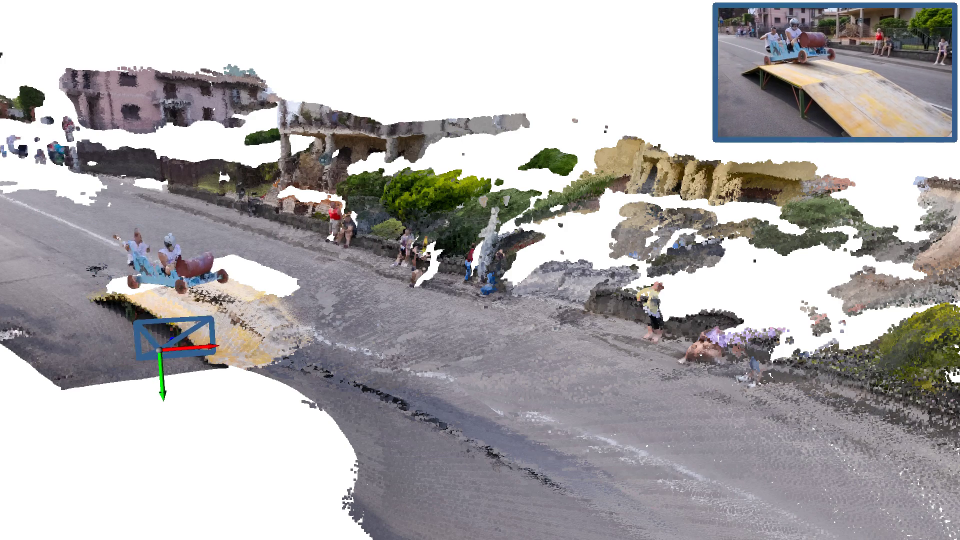} \\
          
          \includegraphics[width=0.33\textwidth, clip]{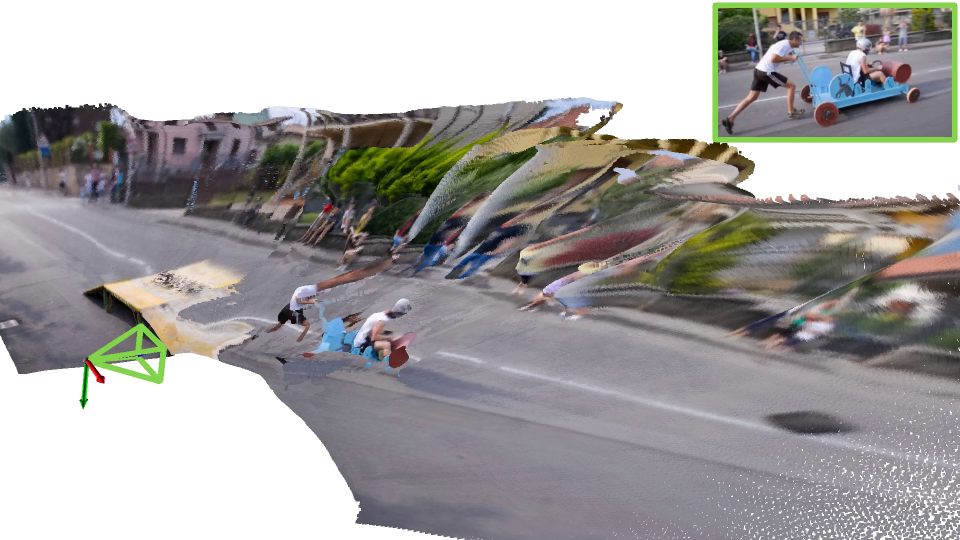} & 
          \includegraphics[width=0.33\textwidth, clip]{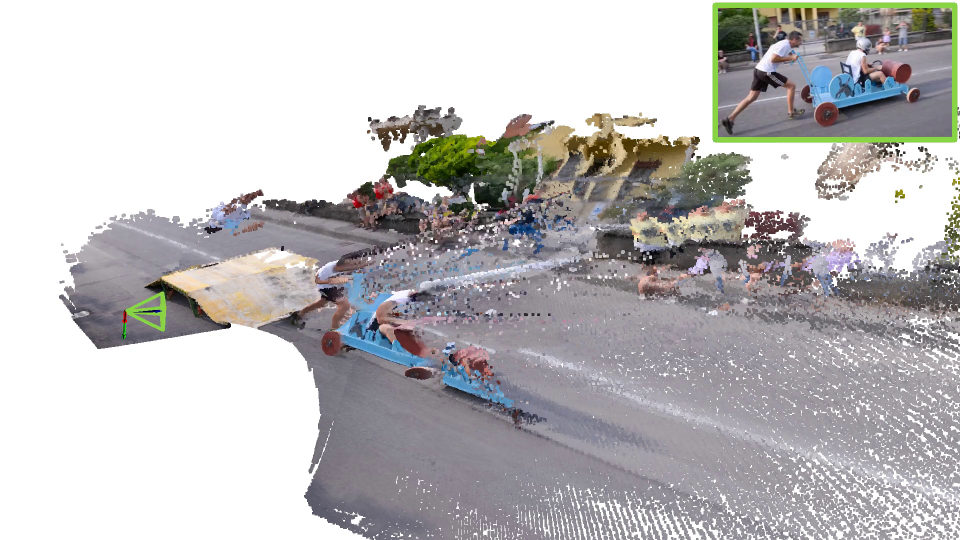} &
          \includegraphics[width=0.33\textwidth, clip]{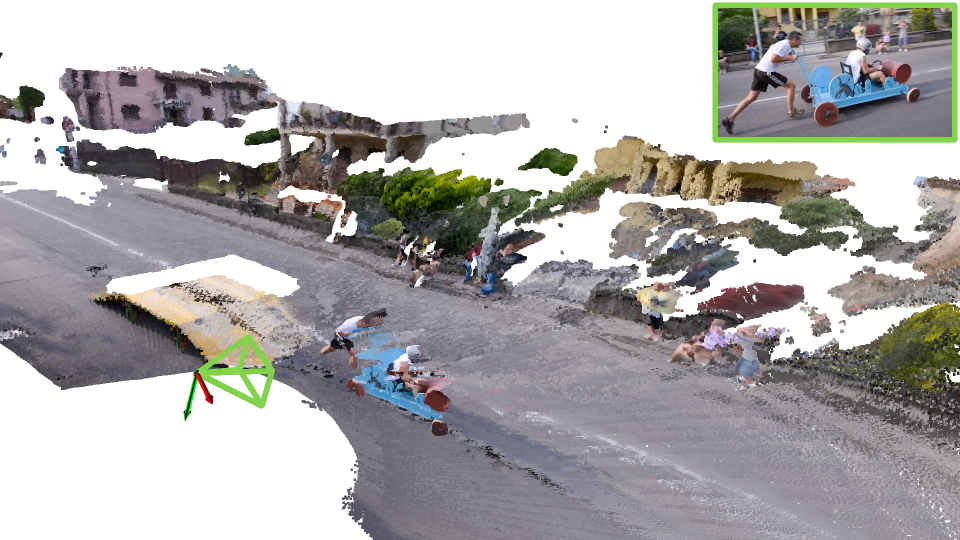} \\
        \hline 
          \multicolumn{3}{c}{\textbf{Car-Roundabout}} \\
              \multicolumn{1}{c}{\textbf{\underline{CasualSAM}}} & \multicolumn{1}{c}{\textbf{\underline{MonST3R}}} & \multicolumn{1}{c}{\textbf{\underline{Uni4D (Ours)}}} \\

          \includegraphics[width=0.33\textwidth, clip]{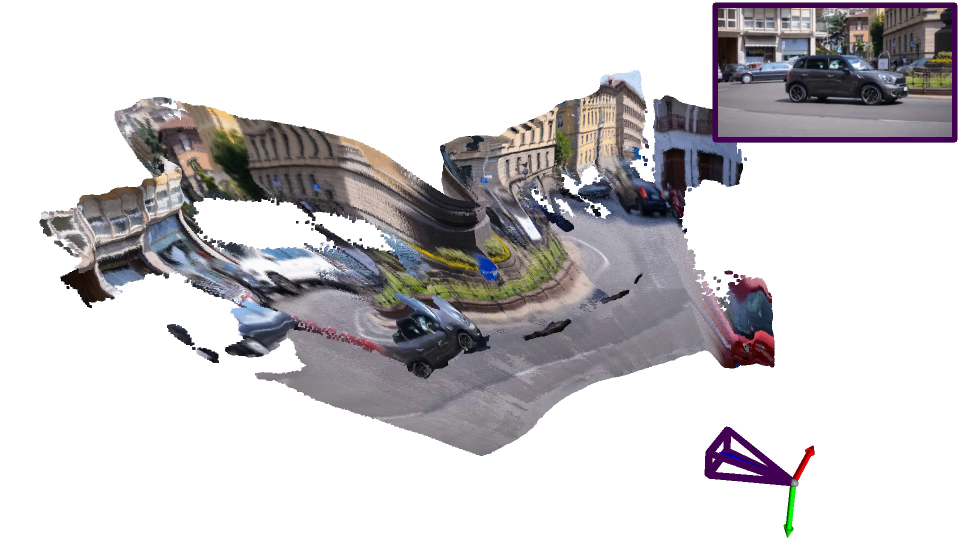} & 
          \includegraphics[width=0.33\textwidth, clip]{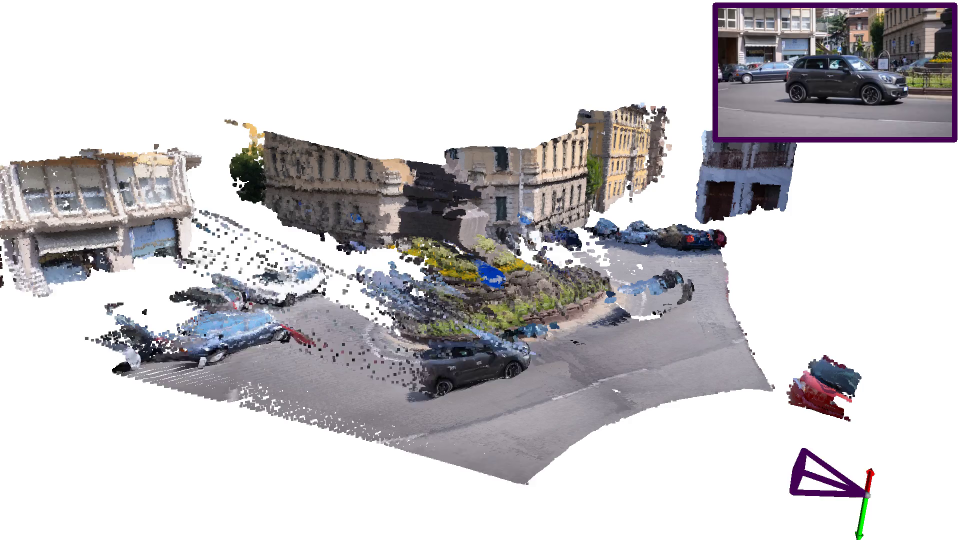} &
          \includegraphics[width=0.33\textwidth, clip]{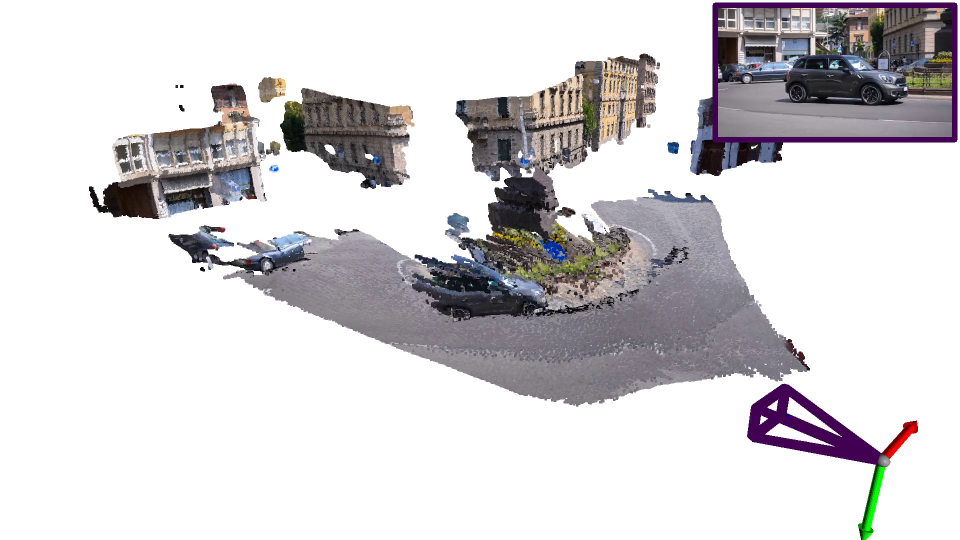} \\
          
          \includegraphics[width=0.33\textwidth, clip]{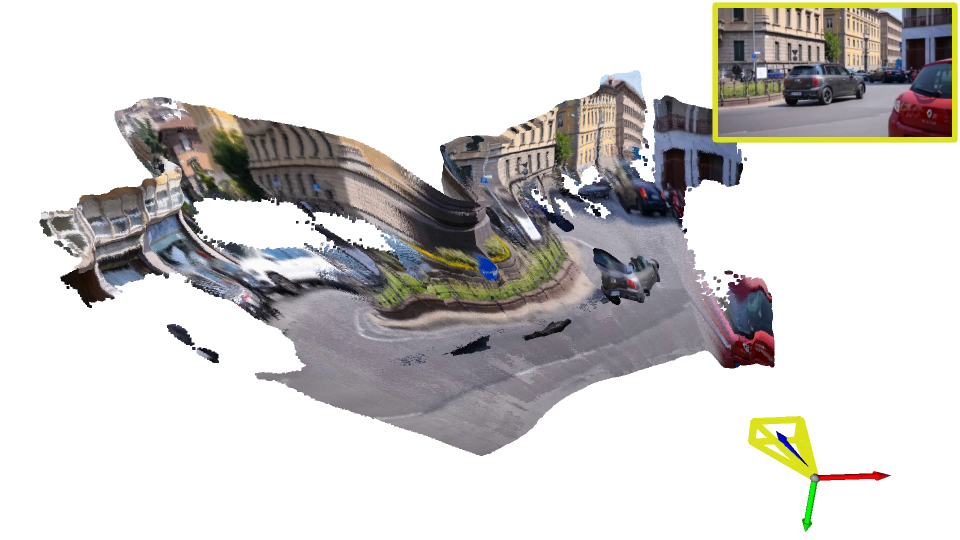} & 
          \includegraphics[width=0.33\textwidth, clip]{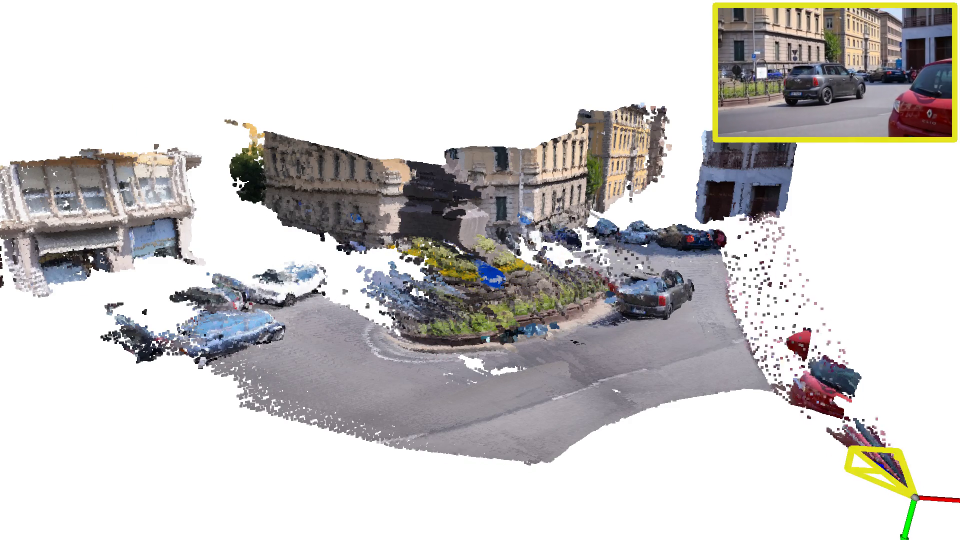} &
          \includegraphics[width=0.33\textwidth, clip]{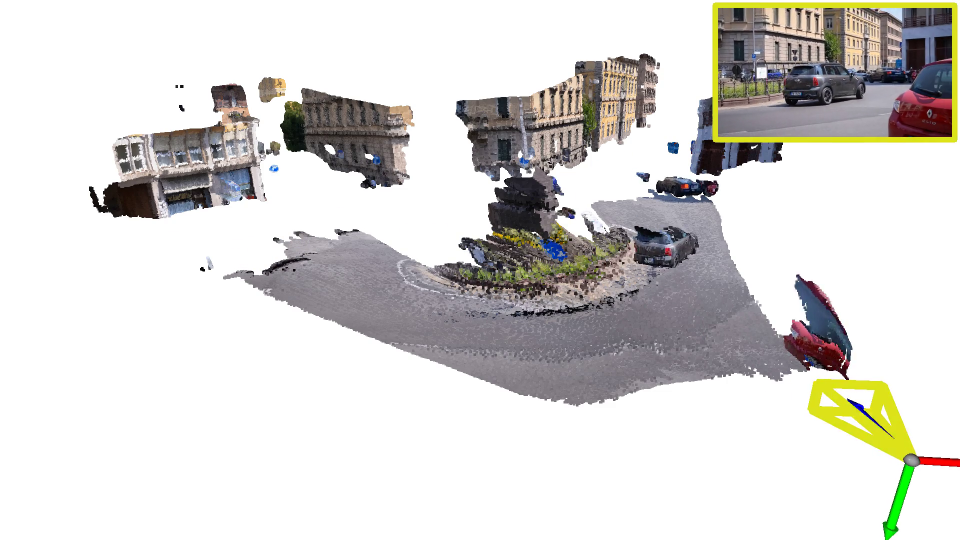} \\
          \hline 
          \multicolumn{3}{c}{\textbf{Snowboard}} \\
              \multicolumn{1}{c}{\textbf{\underline{CasualSAM}}} & \multicolumn{1}{c}{\textbf{\underline{MonST3R}}} & \multicolumn{1}{c}{\textbf{\underline{Uni4D (Ours)}}} \\

          \includegraphics[width=0.33\textwidth, clip]{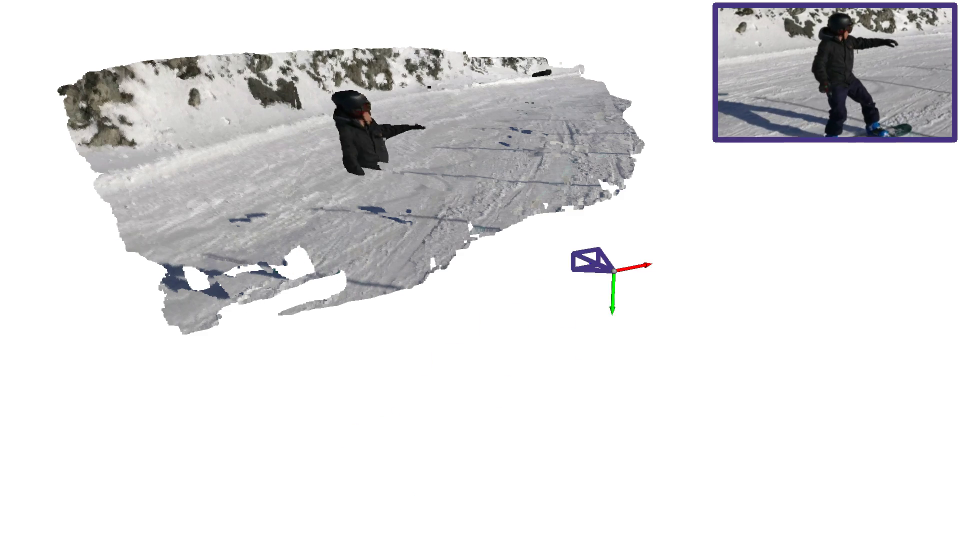} & 
          \includegraphics[width=0.33\textwidth, clip]{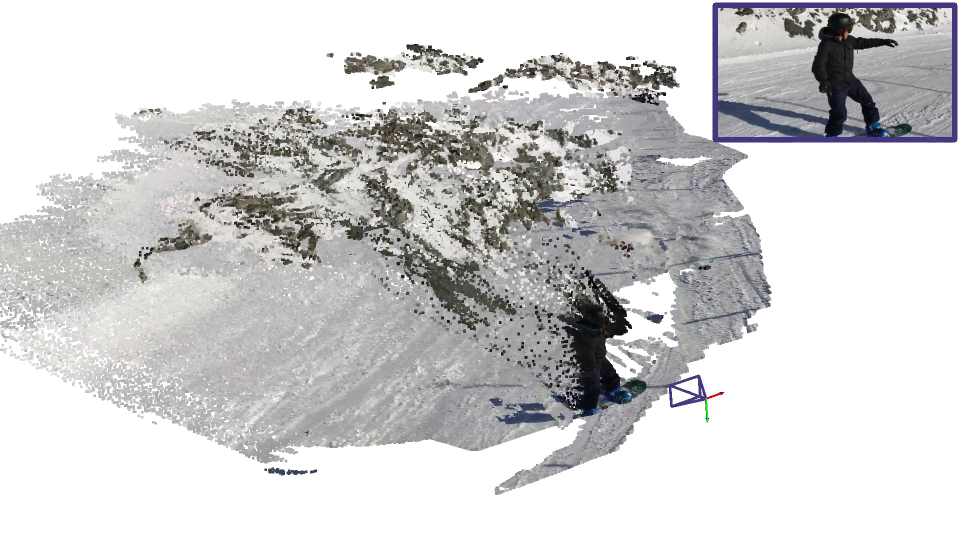} &
          \includegraphics[width=0.33\textwidth, clip]{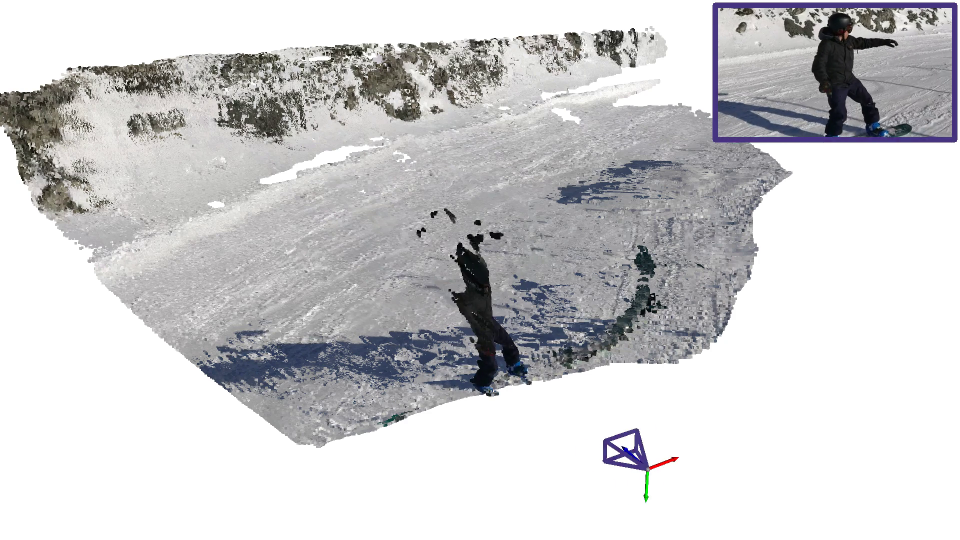} \\
          
          \includegraphics[width=0.33\textwidth, clip]{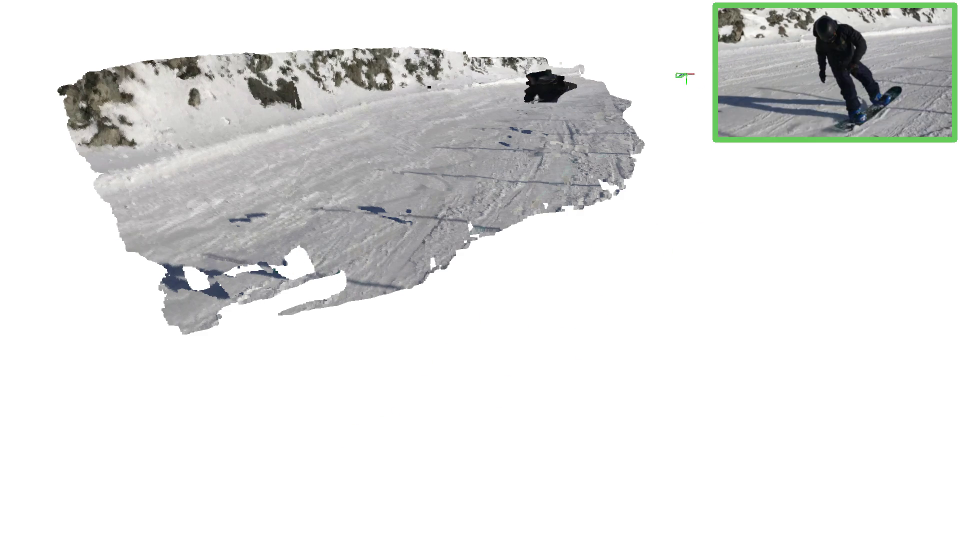} & 
          \includegraphics[width=0.33\textwidth, clip]{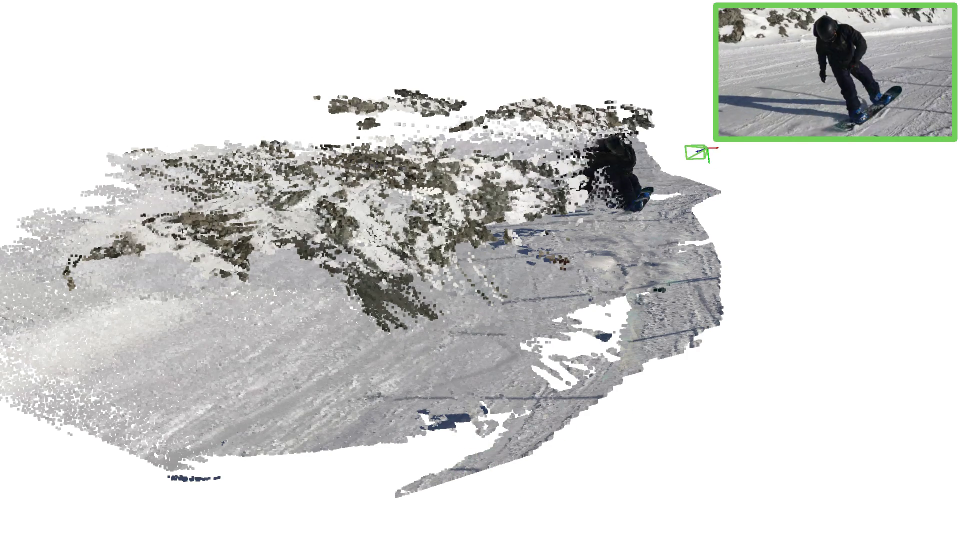} &
          \includegraphics[width=0.33\textwidth, clip]{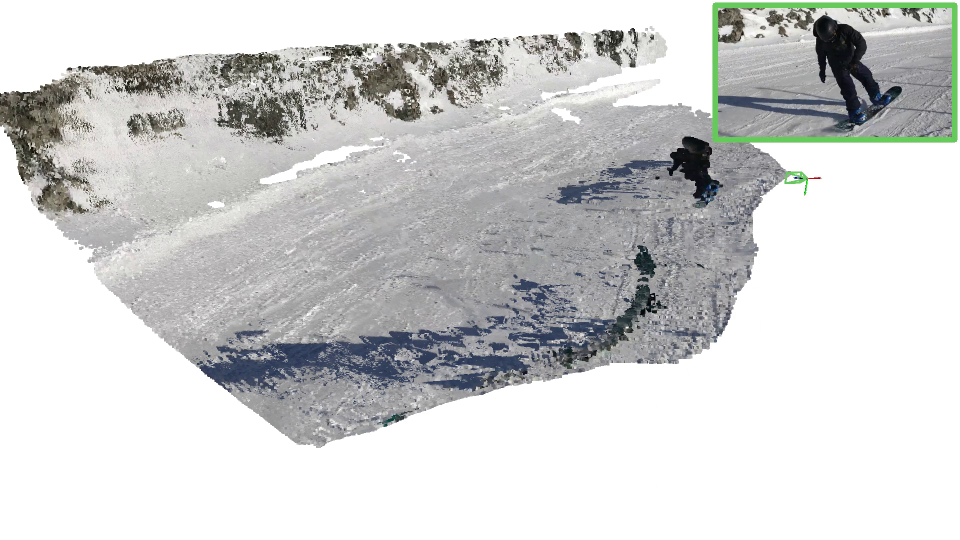} \\
    \end{tabular}
    }
    \vspace{-3mm}
    \captionof{figure}{{\bf Qualitative Results on DAVIS dataset} {We show qualitatively some of our reconstruction results on the DAVIS dataset compared with other baselines. We visualize here two temporally separate frames and their reconstructions. For full reconstruction, please refer to our attached supplementary webpage.}}
    \label{fig:davis_supp_qual}
\end{table*}

\begin{table*}[t]
    \centering
    \setlength\tabcolsep{0.05em} %
    \vspace{-2mm}\resizebox{\textwidth}{!}{
    \begin{tabular}{ccc}

          \hline
          \multicolumn{3}{c}{\textbf{Sintel}} \\
          \multicolumn{1}{c}{\textbf{\underline{CasualSAM}}} & \multicolumn{1}{c}{\textbf{\underline{MonST3R}}} & \multicolumn{1}{c}{\textbf{\underline{Uni4D (Ours)}}} \\ 

          \includegraphics[width=0.33\textwidth, clip]{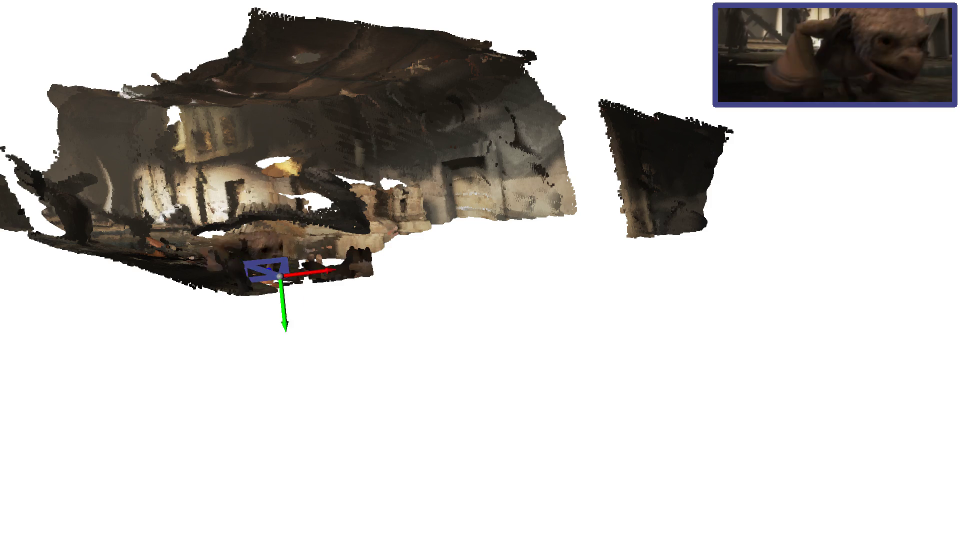} & 
          \includegraphics[width=0.33\textwidth, clip]{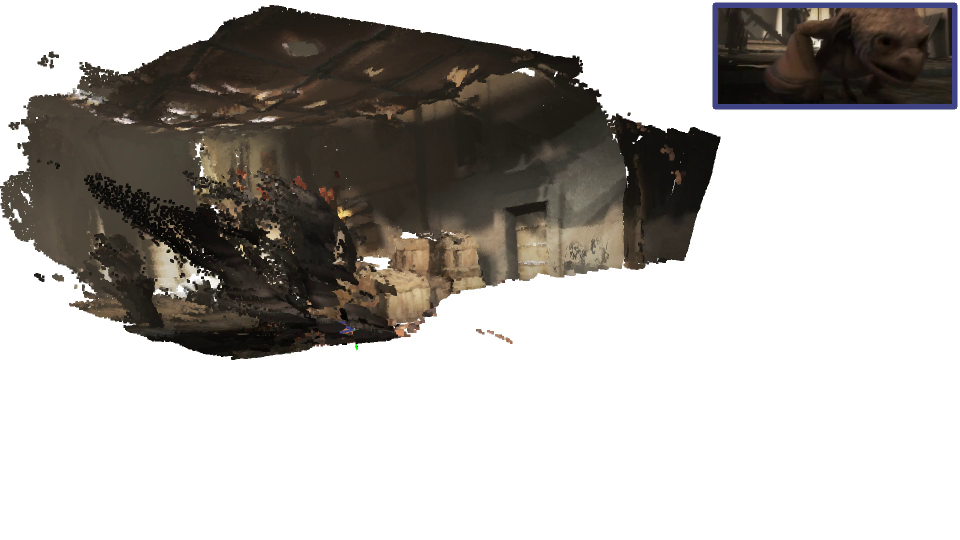} &
          \includegraphics[width=0.33\textwidth, clip]{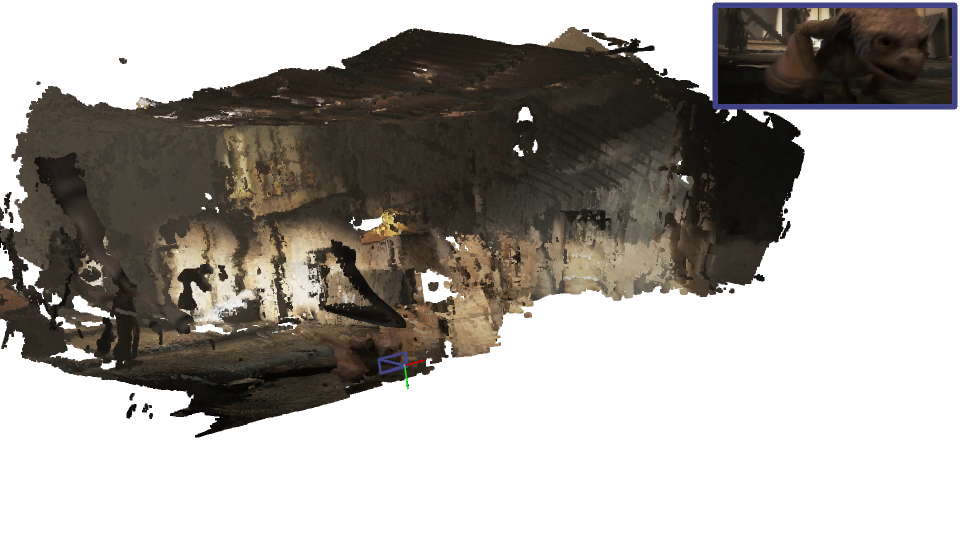} \\
          
          \includegraphics[width=0.33\textwidth, clip]{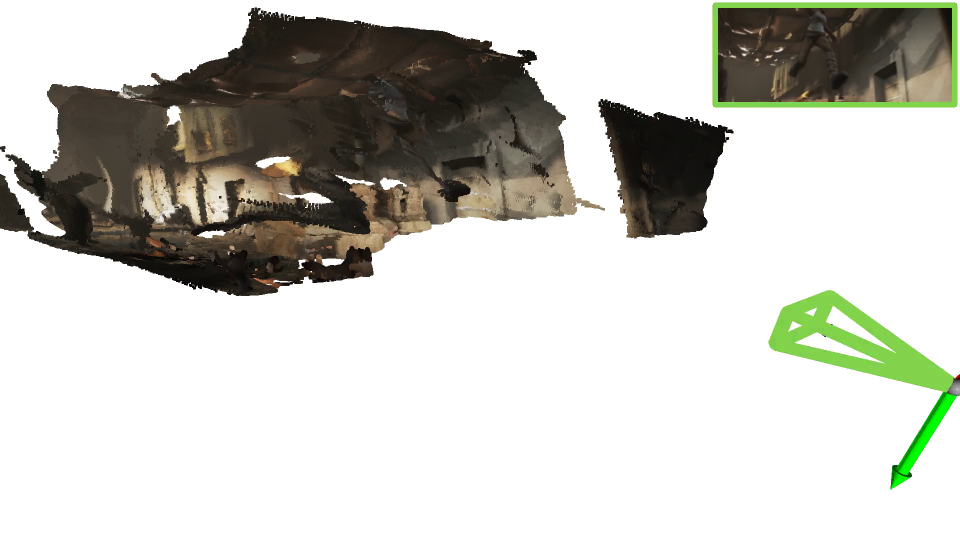} & 
          \includegraphics[width=0.33\textwidth, clip]{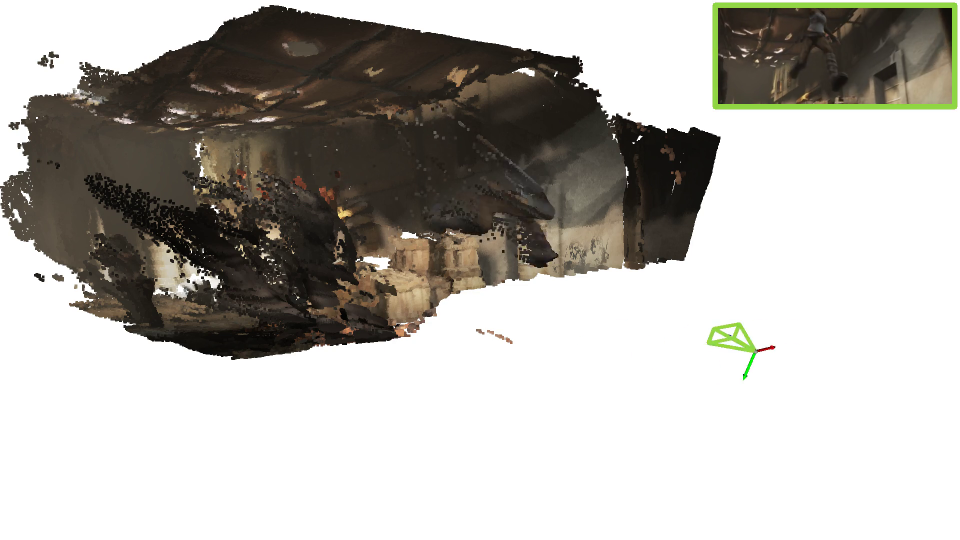} &
          \includegraphics[width=0.33\textwidth, clip]{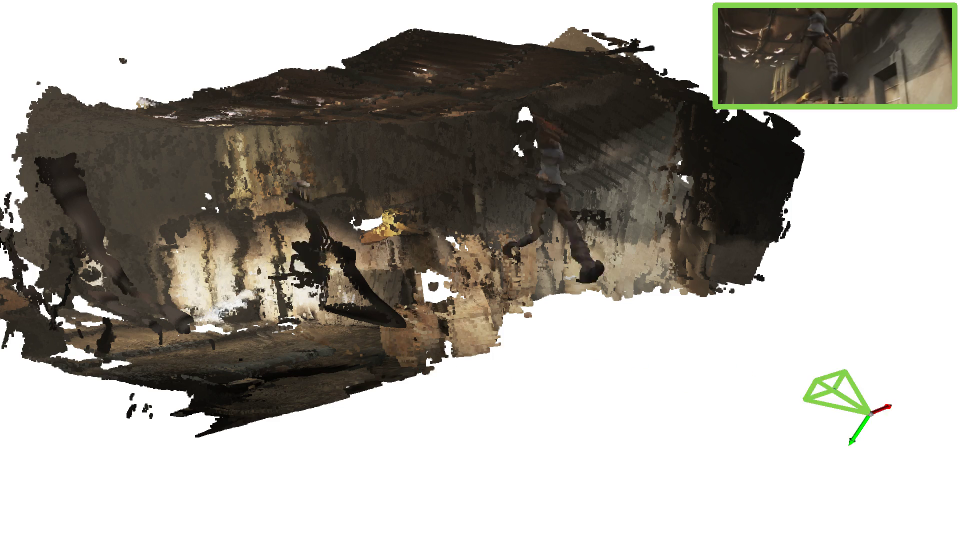} \\

          \hline
          \multicolumn{3}{c}{\textbf{TUM-Dynamics}} \\
              \multicolumn{1}{c}{\textbf{\underline{CasualSAM}}} & \multicolumn{1}{c}{\textbf{\underline{MonST3R}}} & \multicolumn{1}{c}{\textbf{\underline{Uni4D (Ours)}}} \\ 

          \includegraphics[width=0.33\textwidth, clip]{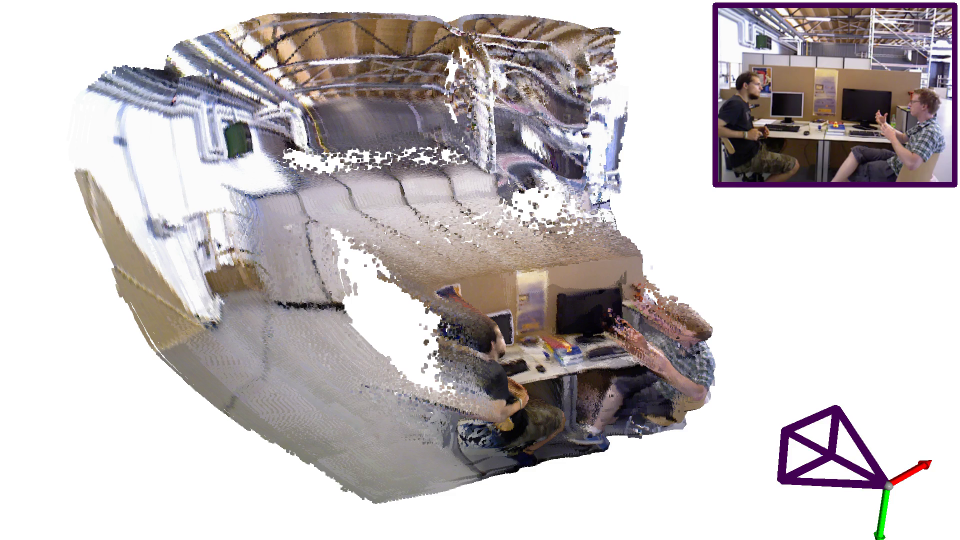} & 
          \includegraphics[width=0.33\textwidth, clip]{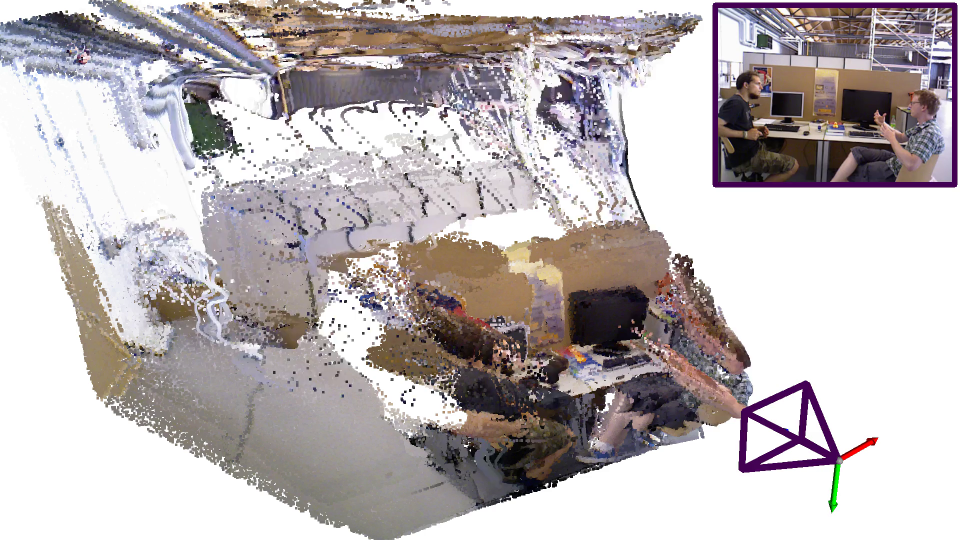} &
          \includegraphics[width=0.33\textwidth, clip]{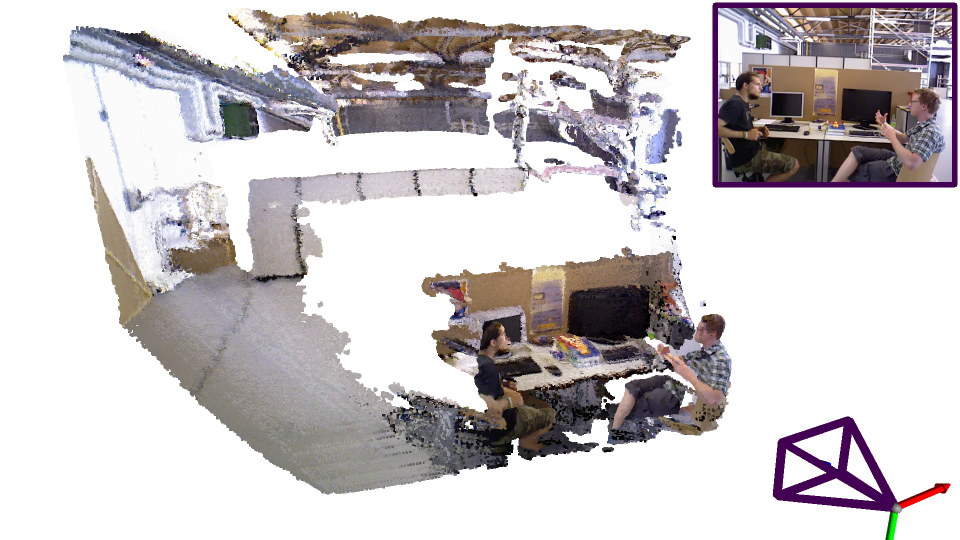} \\
          
          \includegraphics[width=0.33\textwidth, clip]{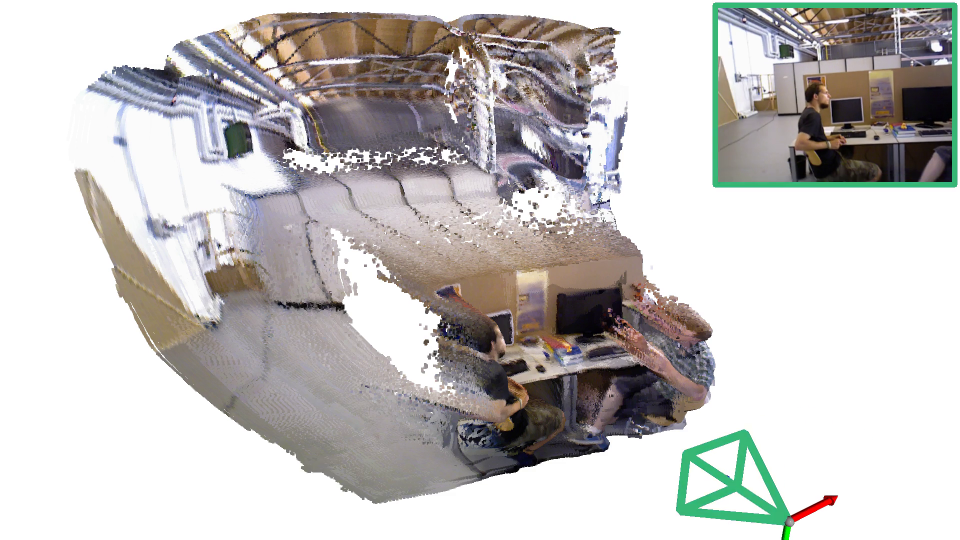} & 
          \includegraphics[width=0.33\textwidth, clip]{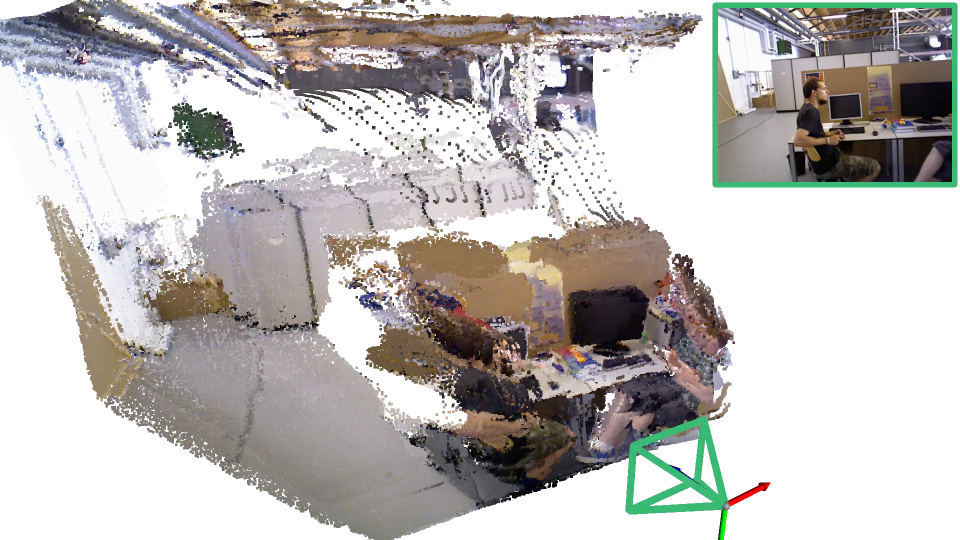} &
          \includegraphics[width=0.33\textwidth, clip]{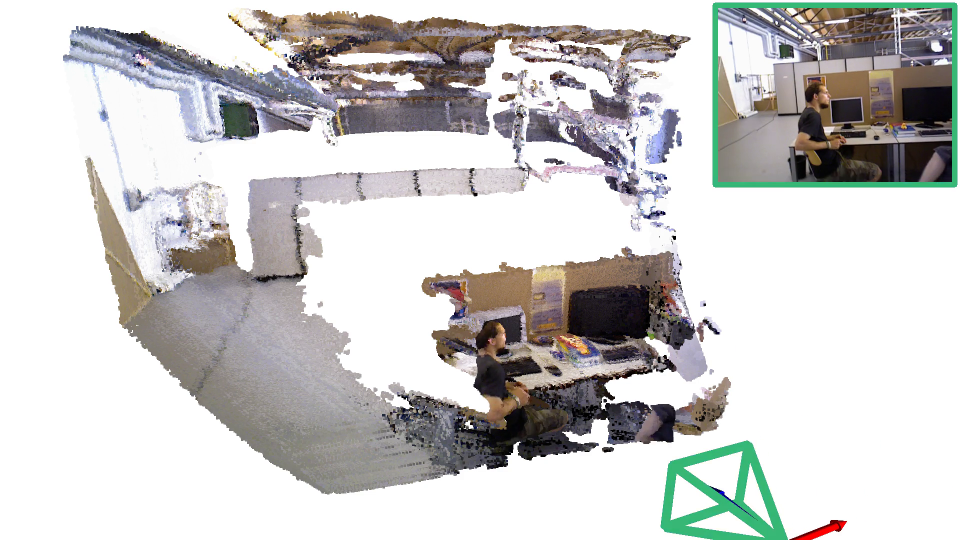} \\

    \end{tabular}
    }
    \vspace{-3mm}
    \captionof{figure}{{\bf Qualitative Results on Sintel and TUM-Dynamics dataset} We show qualitatively some of our reconstruction results on Sintel and TUM-Dynamics dataset compared with other baselines. We visualize here 2 temporally separate frames and their reconstructions. For full reconstruction, please refer to our attached supplementary webpage.}
    \label{fig:sintel_tumd_supp_qual}
\end{table*}

\begin{table*}[t]
    \centering
    \setlength\tabcolsep{0.05em} %
    \vspace{-2mm}\resizebox{\textwidth}{!}{
    \begin{tabular}{ccc}

          \hline
          \multicolumn{3}{c}{\textbf{Bonn}} \\
          \multicolumn{1}{c}{\textbf{\underline{CasualSAM}}} & \multicolumn{1}{c}{\textbf{\underline{MonST3R}}} & \multicolumn{1}{c}{\textbf{\underline{Uni4D (Ours)}}} \\ 
          \includegraphics[width=0.33\textwidth, clip]{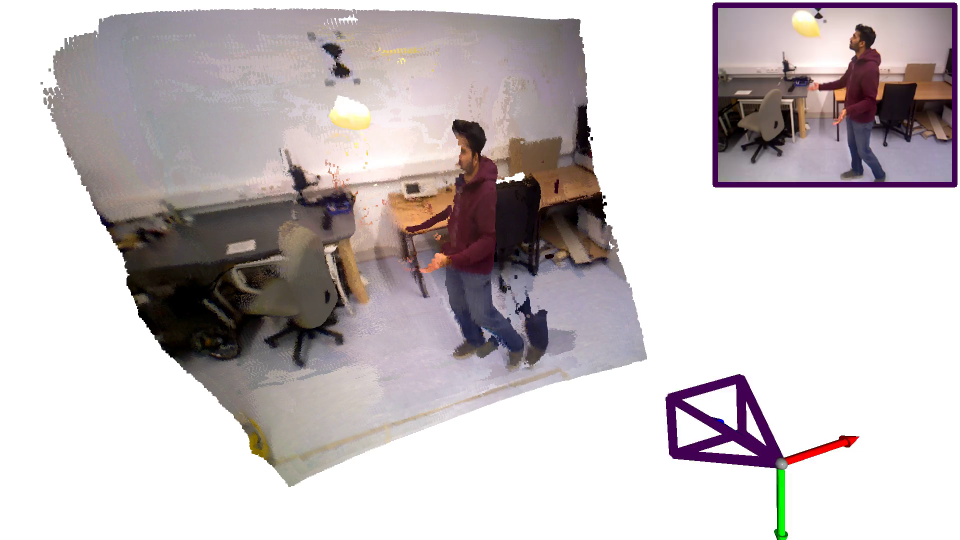} & 
          \includegraphics[width=0.33\textwidth, clip]{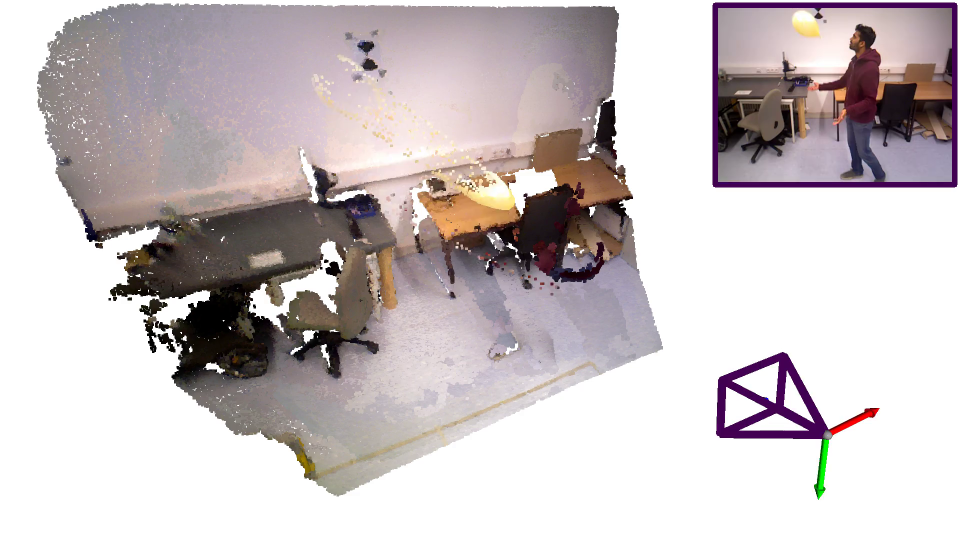} &
          \includegraphics[width=0.33\textwidth, clip]{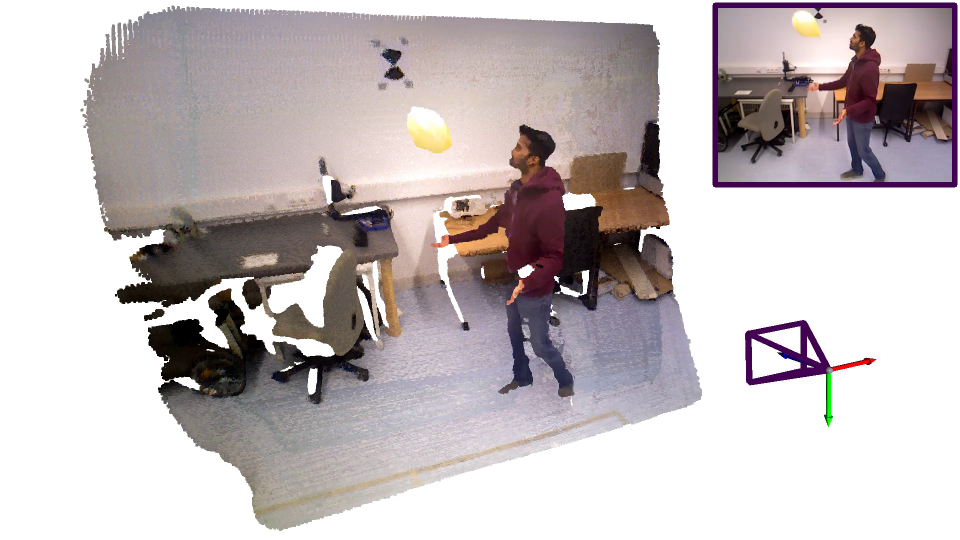} \\
          
          \includegraphics[width=0.33\textwidth, clip]{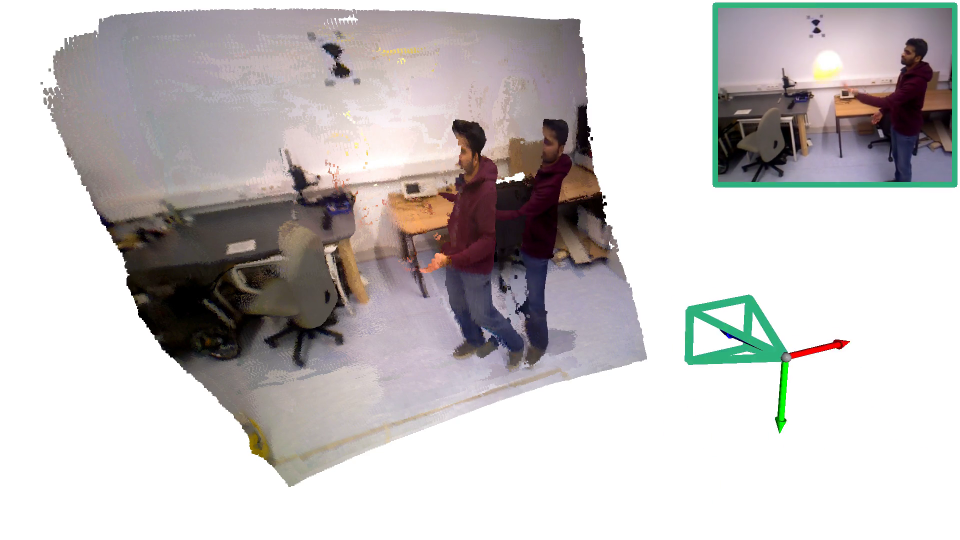} & 
          \includegraphics[width=0.33\textwidth, clip]{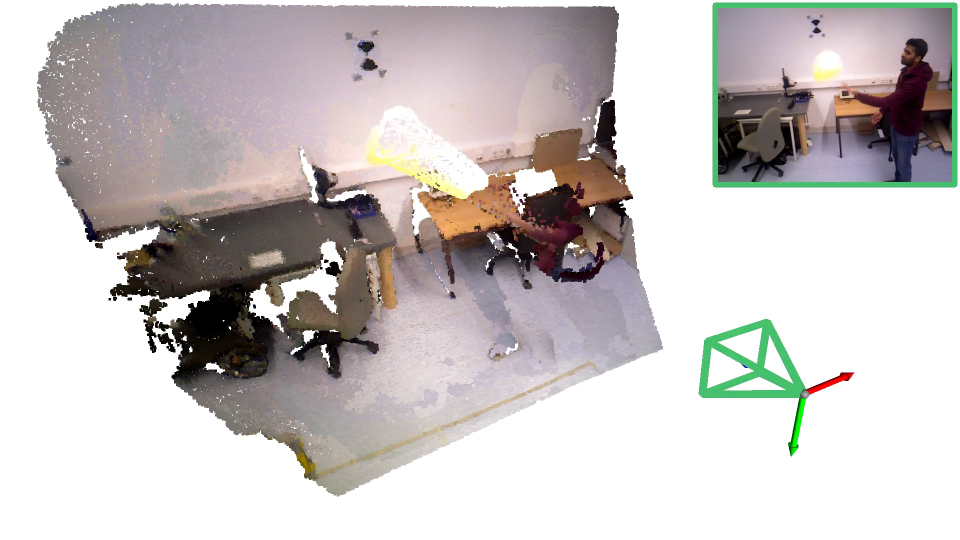} &
          \includegraphics[width=0.33\textwidth, clip]{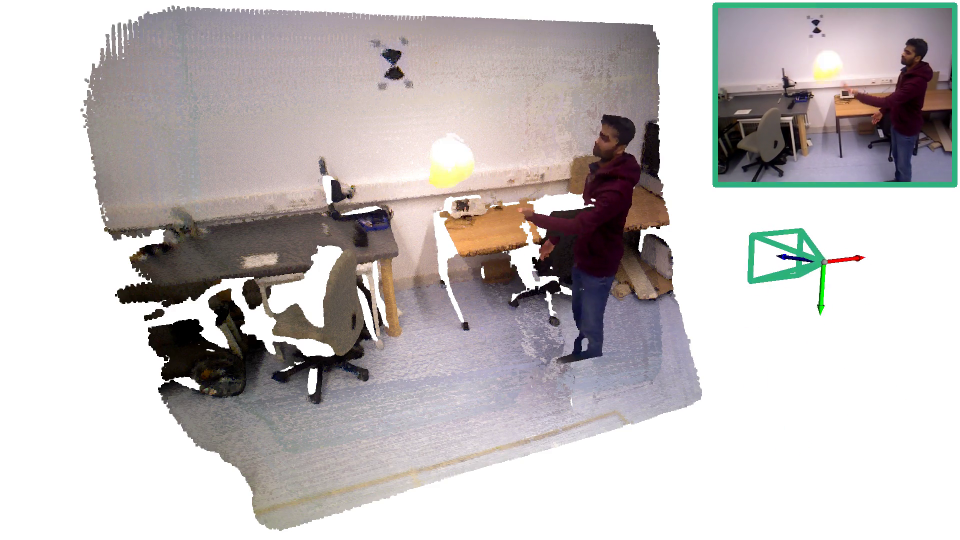} \\

            \hline 
            \multicolumn{3}{c}{\textbf{KITTI}} \\
              \multicolumn{1}{c}{\textbf{\underline{CasualSAM}}} & \multicolumn{1}{c}{\textbf{\underline{MonST3R}}} & \multicolumn{1}{c}{\textbf{\underline{Uni4D (Ours)}}} \\ 
          \includegraphics[width=0.33\textwidth, clip]{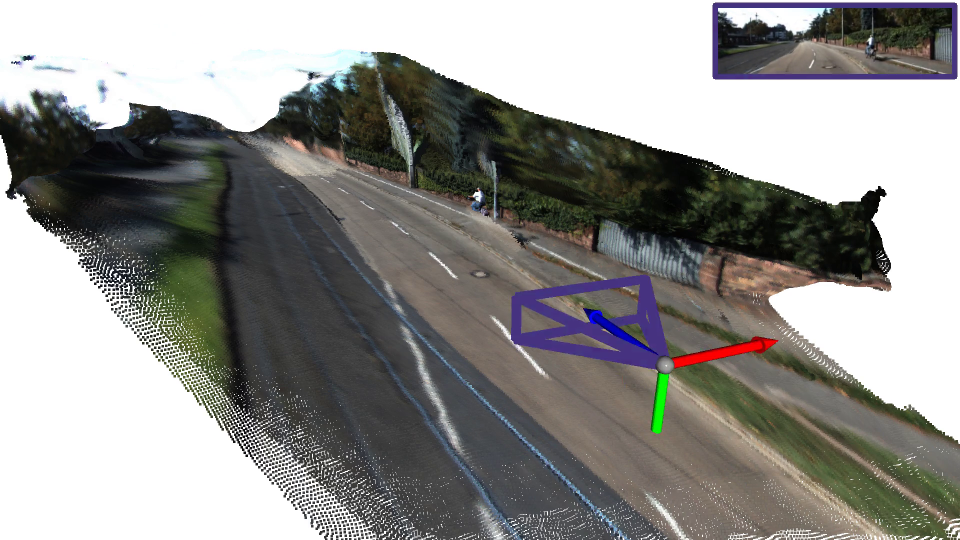} & 
          \includegraphics[width=0.33\textwidth, clip]{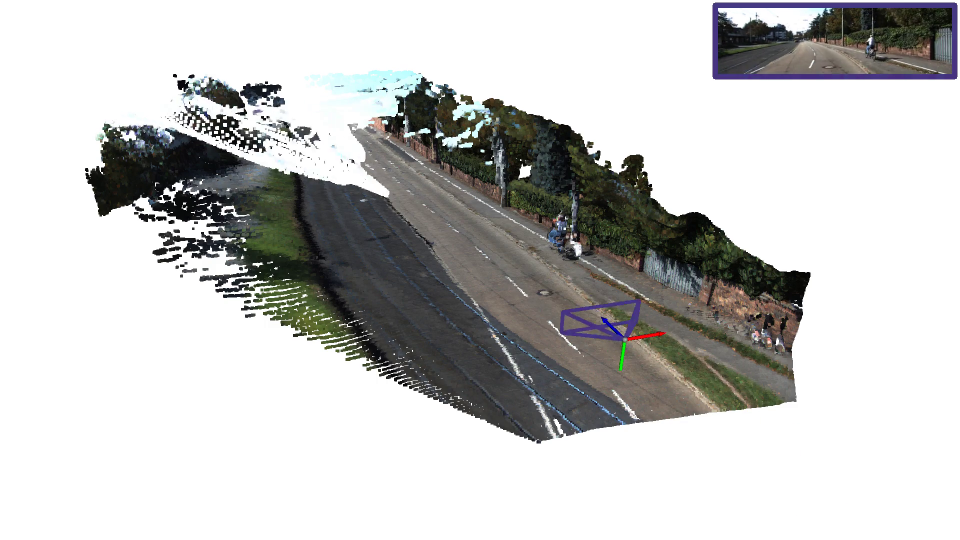} &
          \includegraphics[width=0.33\textwidth, clip]{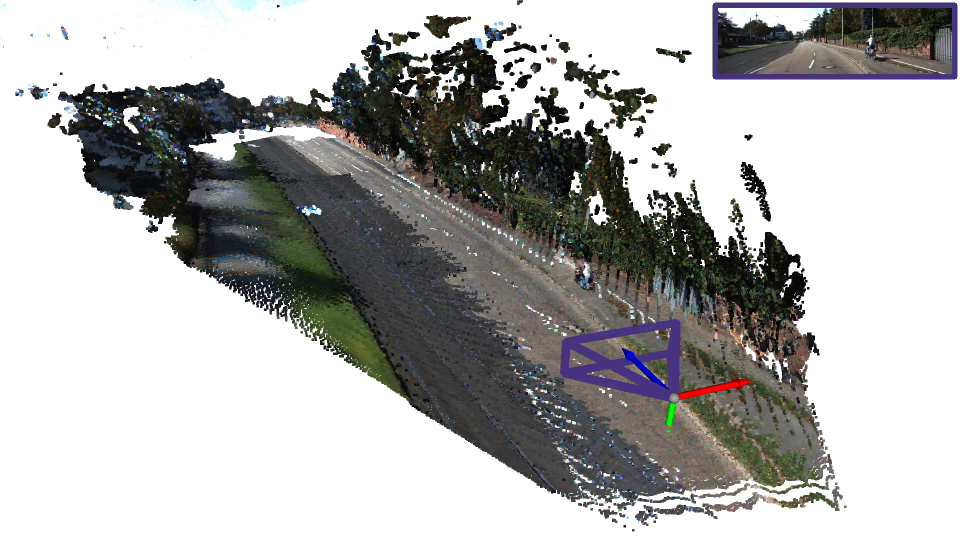} \\
          
          \includegraphics[width=0.33\textwidth, clip]{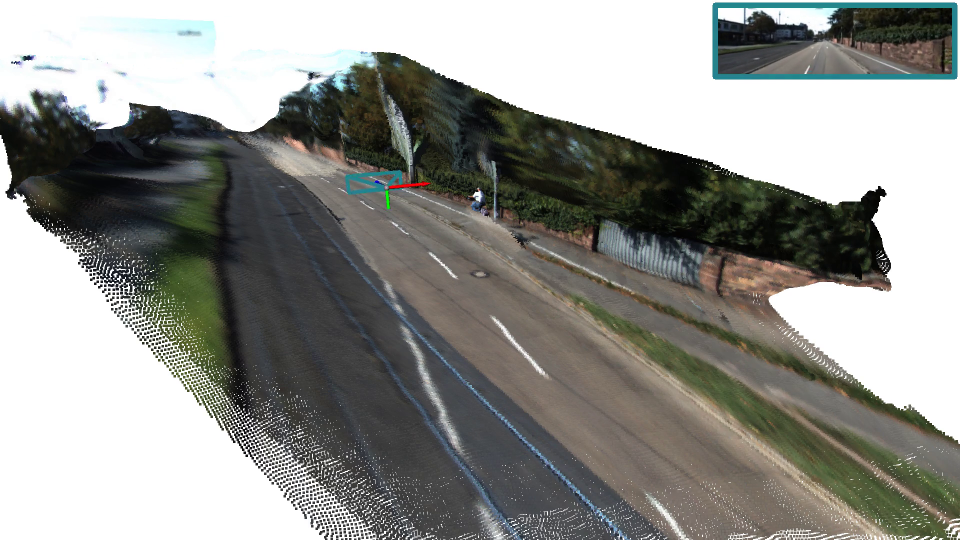} & 
          \includegraphics[width=0.33\textwidth, clip]{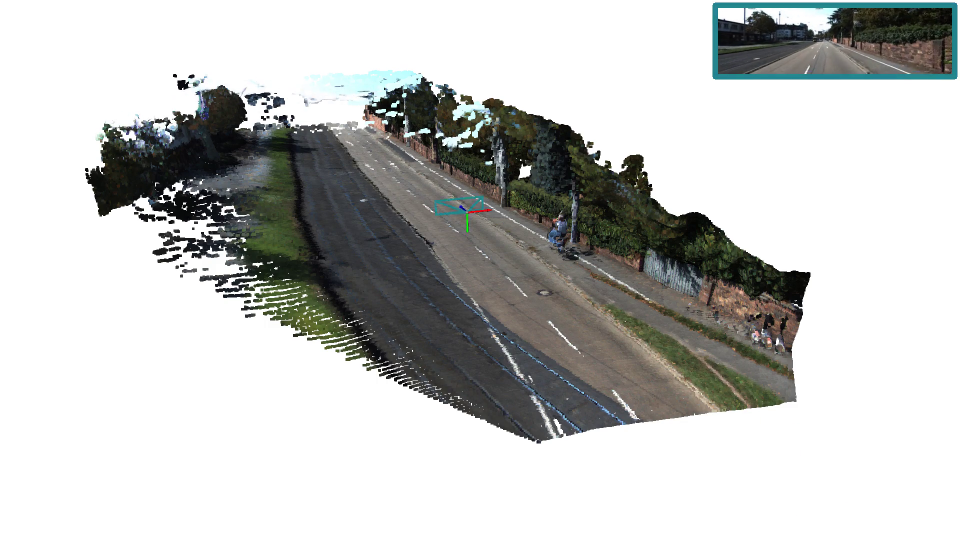} &
          \includegraphics[width=0.33\textwidth,, clip]{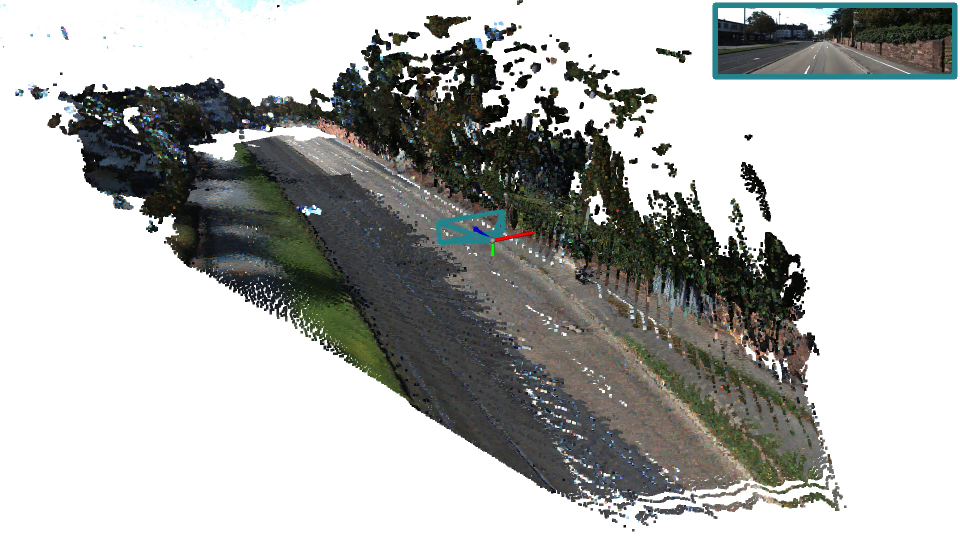} \\

    \end{tabular}
    }
    \vspace{-3mm}
    \captionof{figure}{{\bf Qualitative Results on Bonn and KITTI dataset}We show qualitatively some of our reconstruction results on Bonn and KITTI dataset compared with other baselines. We visualize here 2 temporally separate frames and their reconstructions. For full reconstruction, please refer to our attached supplementary webpage.}
    \label{fig:bon_kitti_supp_qual}
\end{table*}

\begin{table*}[t]
    \centering
    \setlength\tabcolsep{0.05em} %
    \vspace{-2mm}\resizebox{\textwidth}{!}{
    \begin{tabular}{ccc}   
        \hline \\
          \includegraphics[width=0.33\textwidth, clip]{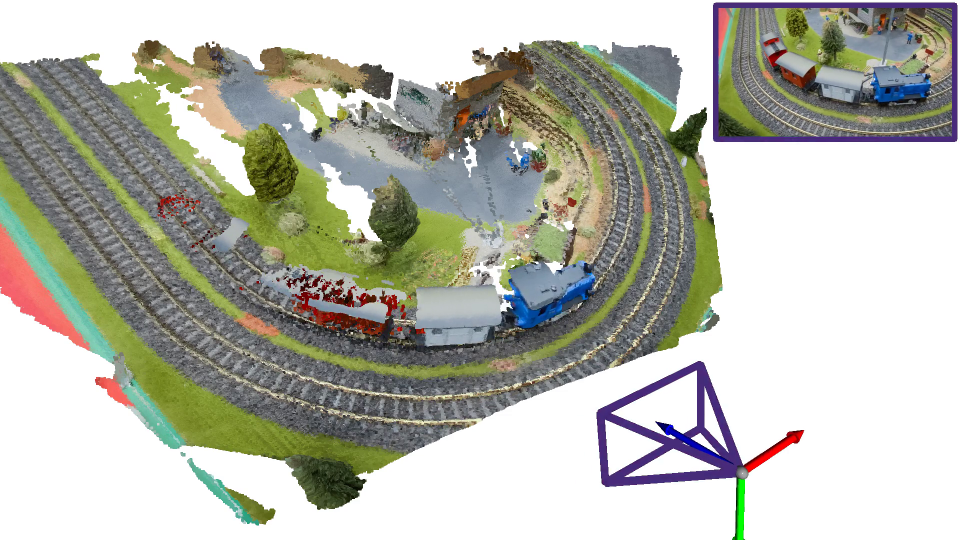} & 
          \includegraphics[width=0.33\textwidth, clip]{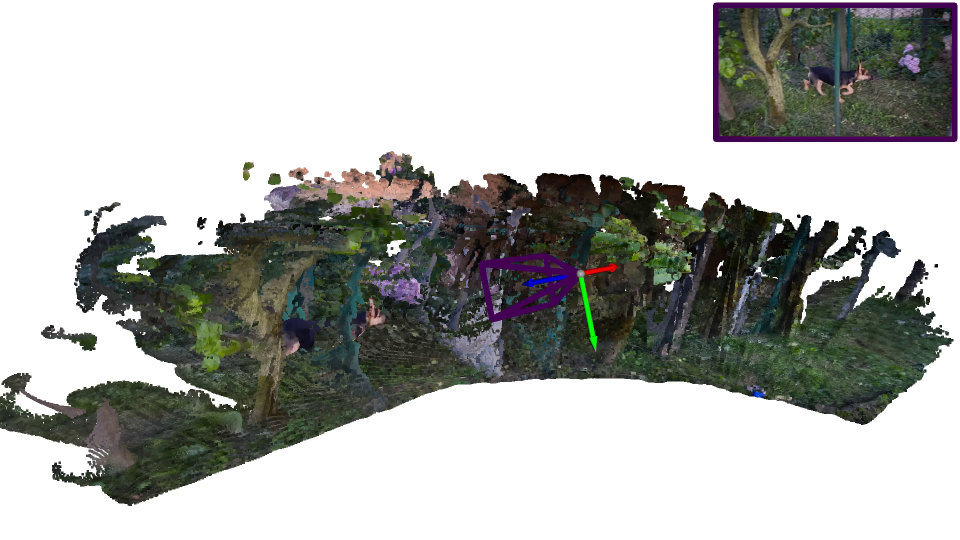} &
          \includegraphics[width=0.33\textwidth, clip]{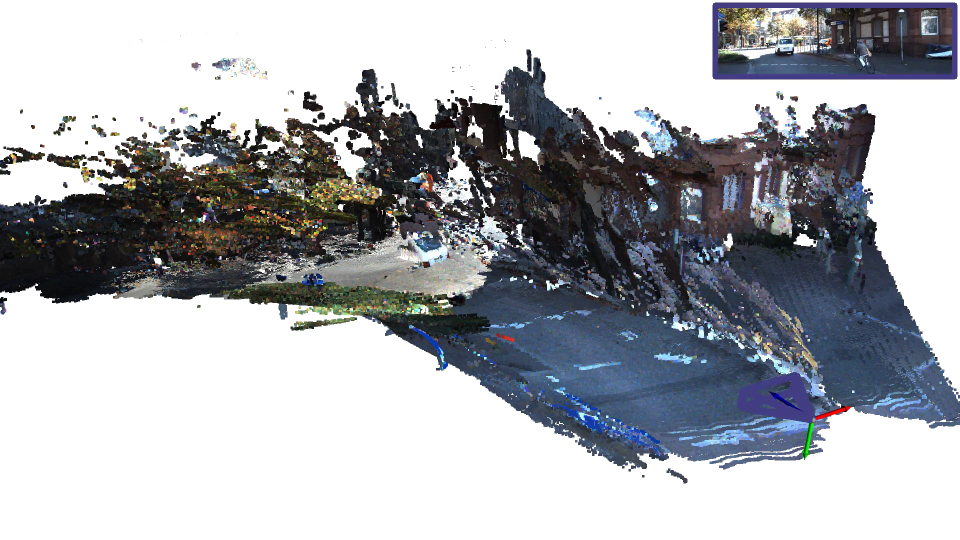} \\
          
          \includegraphics[width=0.33\textwidth, clip]{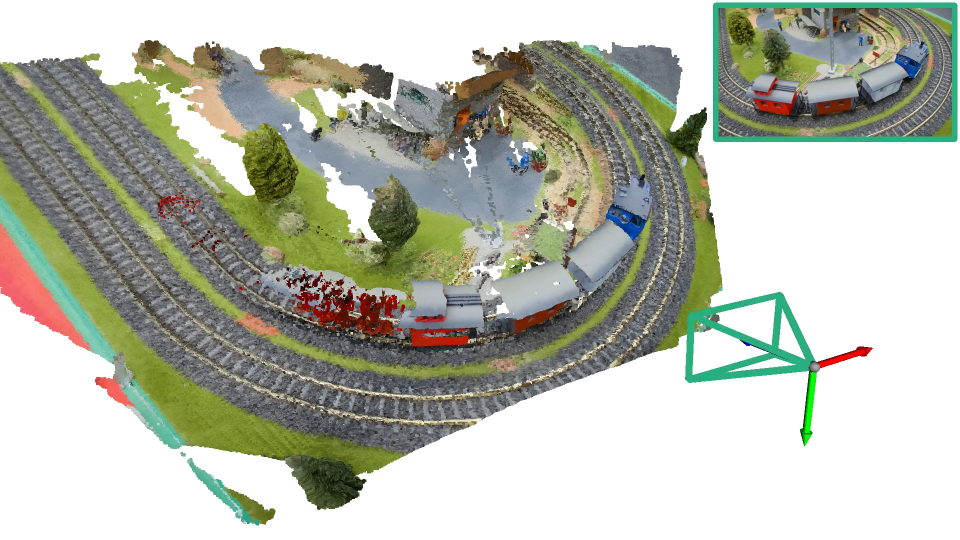} & 
          \includegraphics[width=0.33\textwidth, clip]{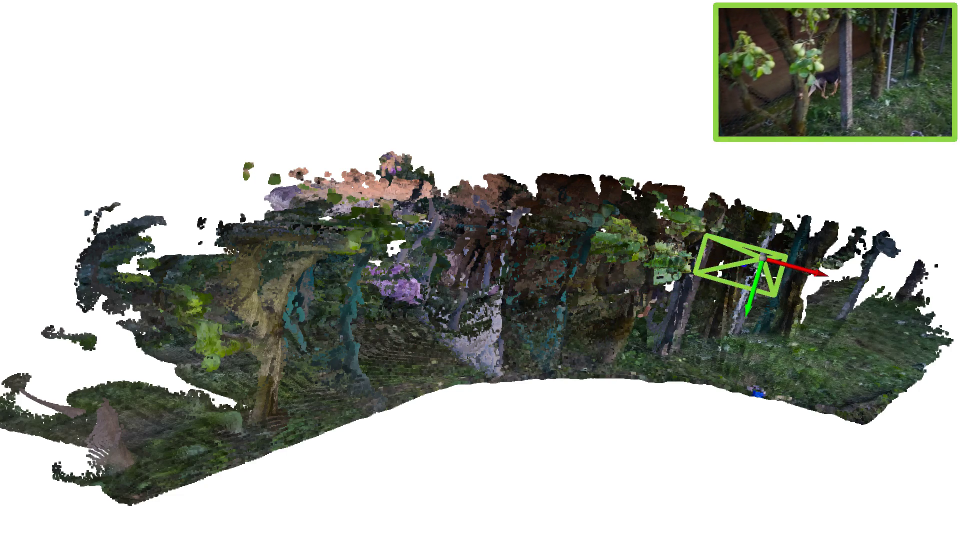} &
          \includegraphics[width=0.33\textwidth, clip]{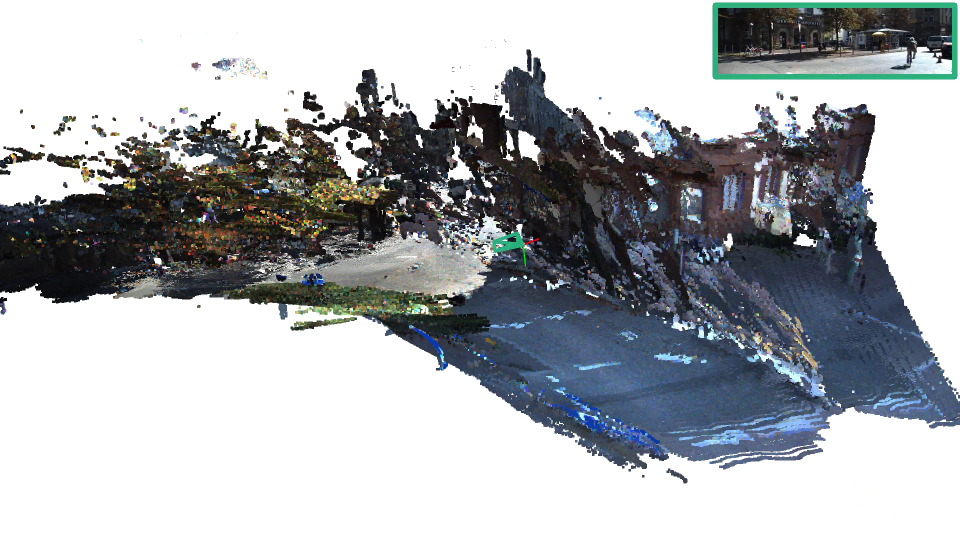} \\
          \multicolumn{1}{c}{\textbf{Incomplete Dynamic Mask}} & \multicolumn{1}{c}{\textbf{Trailing pixels around thin structures}} & \multicolumn{1}{c}{\textbf{Poor Camera Pose}} \\
          \hline \\
          
          \includegraphics[width=0.33\textwidth, clip]{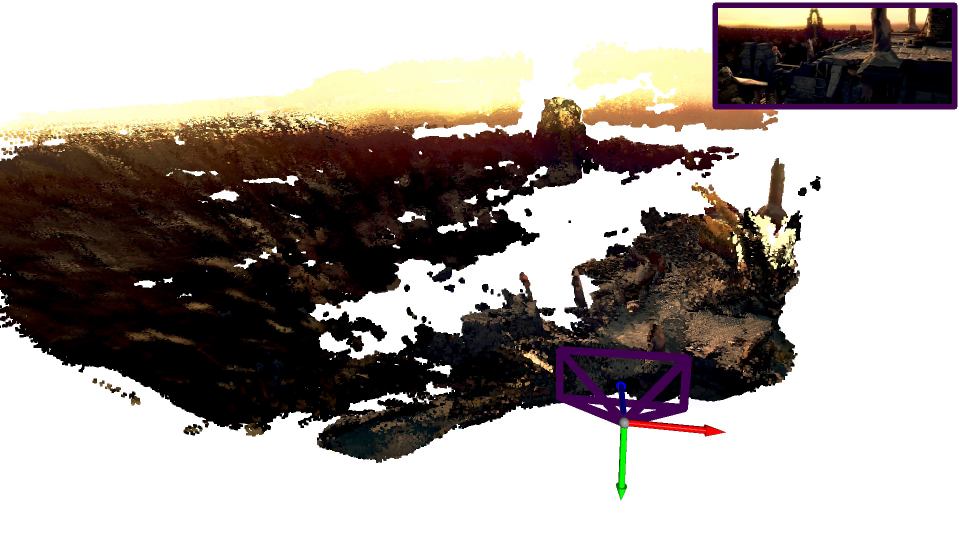} & 
          \includegraphics[width=0.33\textwidth, clip]{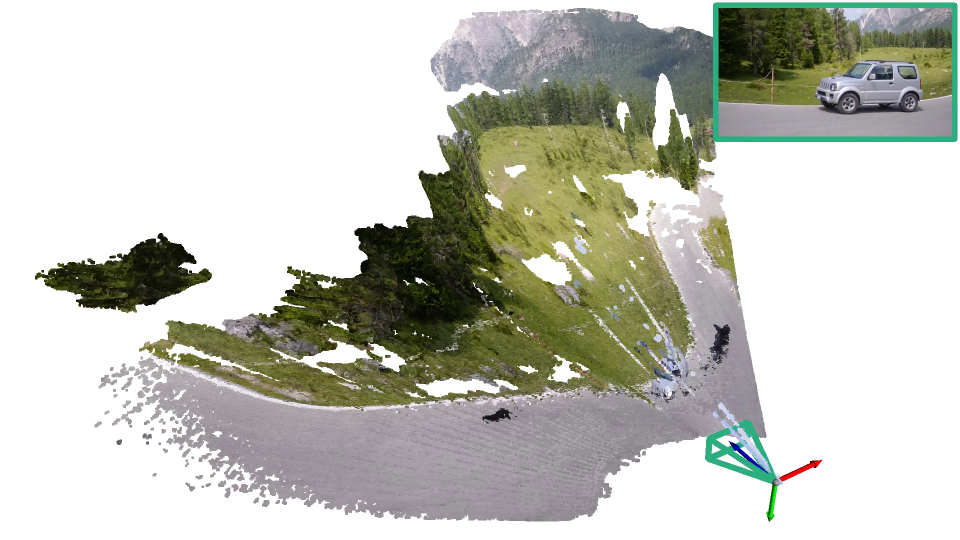} &
          \includegraphics[width=0.33\textwidth, clip]{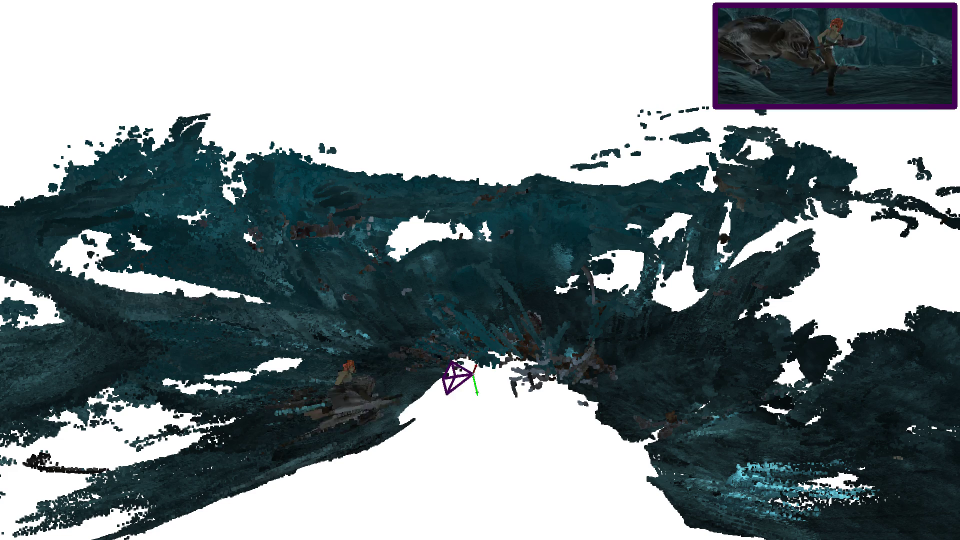} \\
          
          \includegraphics[width=0.33\textwidth, clip]{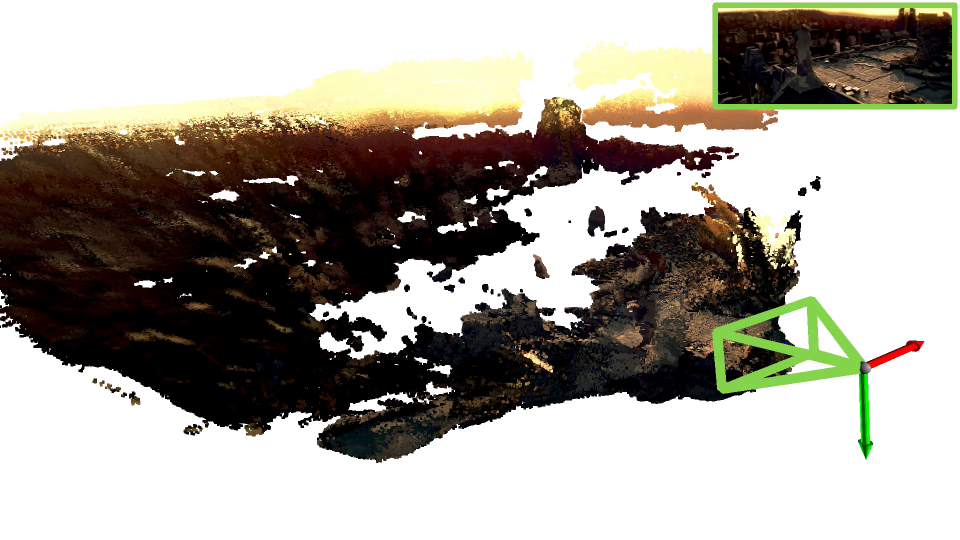} & 
          \includegraphics[width=0.33\textwidth, clip]{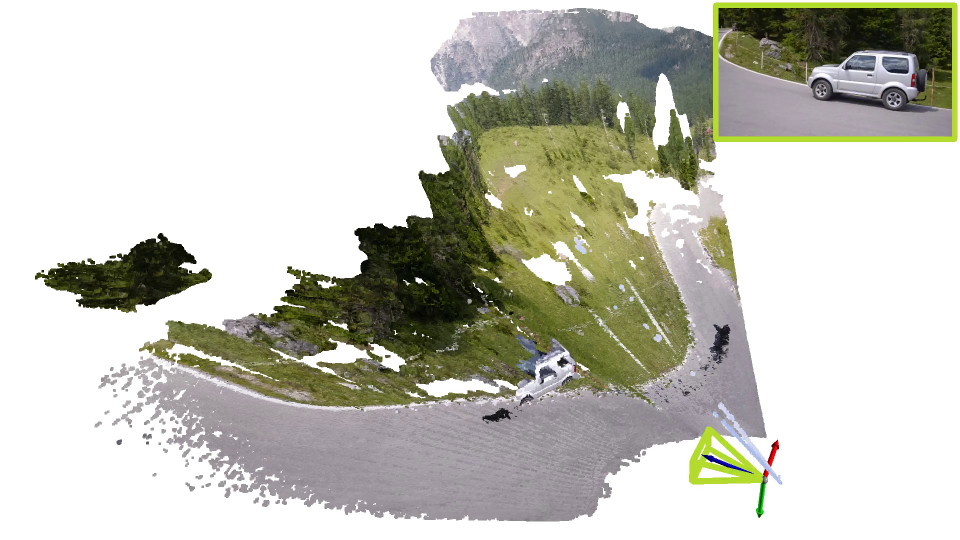} &
          \includegraphics[width=0.33\textwidth, clip]{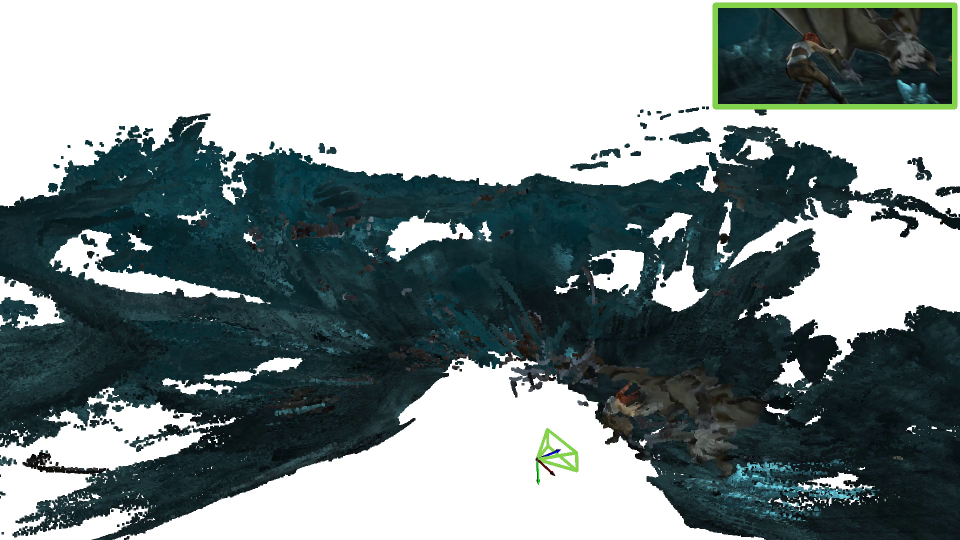} \\
          
          \multicolumn{1}{c}{\textbf{Incorrect Dynamic Masks}} & \multicolumn{1}{c}{\parbox{0.33\textwidth}{\centering\textbf{Trailing pixels around dynamic objects + \\ Incorrect Dynamic Trajectory}}} & \multicolumn{1}{c}{\textbf{Poor Camera Pose + Dynamic Masks}} \\
          \hline \\
    \end{tabular}
    }
    \vspace{-3mm}
    \captionof{figure}{{\bf Failure Cases} We visualize several failure cases of Uni4D on various datasets. We visualize here 2 temporally separate frames and their reconstructions. For full reconstruction, please refer to our attached supplementary webpage.}
    \label{fig:failure_cases}
\end{table*}

\end{document}